\documentclass{article}

\PassOptionsToPackage{compress}{natbib}

\usepackage[preprint]{main}
\hypersetup{pdfsubject={},pdfauthor={Zhuoyuan Wu, Jun Gao},pdftitle={OSCAR: Omni-Embodiment Action-Conditioned World Model for Robotics},pdfkeywords={world models, cross-embodiment robot learning, policy evaluation}}

\usepackage[utf8]{inputenc}
\usepackage[T1]{fontenc}
\usepackage{booktabs}
\usepackage{footnote}
\makesavenoteenv{table}
\usepackage{amsfonts}
\usepackage{amsmath}
\usepackage{amssymb}
\usepackage{graphicx}
\usepackage{tikz}      
\usetikzlibrary{calc,arrows.meta}
\usepackage{wrapfig} 
\usepackage{nicefrac}
\usepackage{microtype}
\usepackage{xcolor}
\usepackage{array}
\usepackage{pifont}    
\usepackage{colortbl}  
\usepackage{multirow} 
\newcommand{\cmark}{\textcolor{green!55!black}{\ding{51}}}
\newcommand{\xmark}{\textcolor{red!75!black}{\ding{55}}}

\usepackage{needspace}
\usepackage[most]{tcolorbox}
\tcbuselibrary{breakable}

\definecolor{oscarcolor}{HTML}{B72025}
\newcommand{\methodname}{\mbox{OSCAR}}
\newcommand{\oscarinit}[1]{\textcolor{oscarcolor}{#1}}

\title{\oscarinit{\methodname{}}: Omni-Embodiment Action-Conditioned World Model for Robotics}

\author{
  Zhuoyuan Wu\textsuperscript{1}\qquad
  Jun Gao\textsuperscript{2,3}\\
  \textsuperscript{1}Peking University\quad
  \textsuperscript{2}University of Michigan \quad
  \textsuperscript{3}NVIDIA
}

\begin{document}
\maketitle

\begin{figure}[h]
  \centering
  \begin{minipage}[c]{0.54\linewidth}
    \centering
    \includegraphics[width=\linewidth]{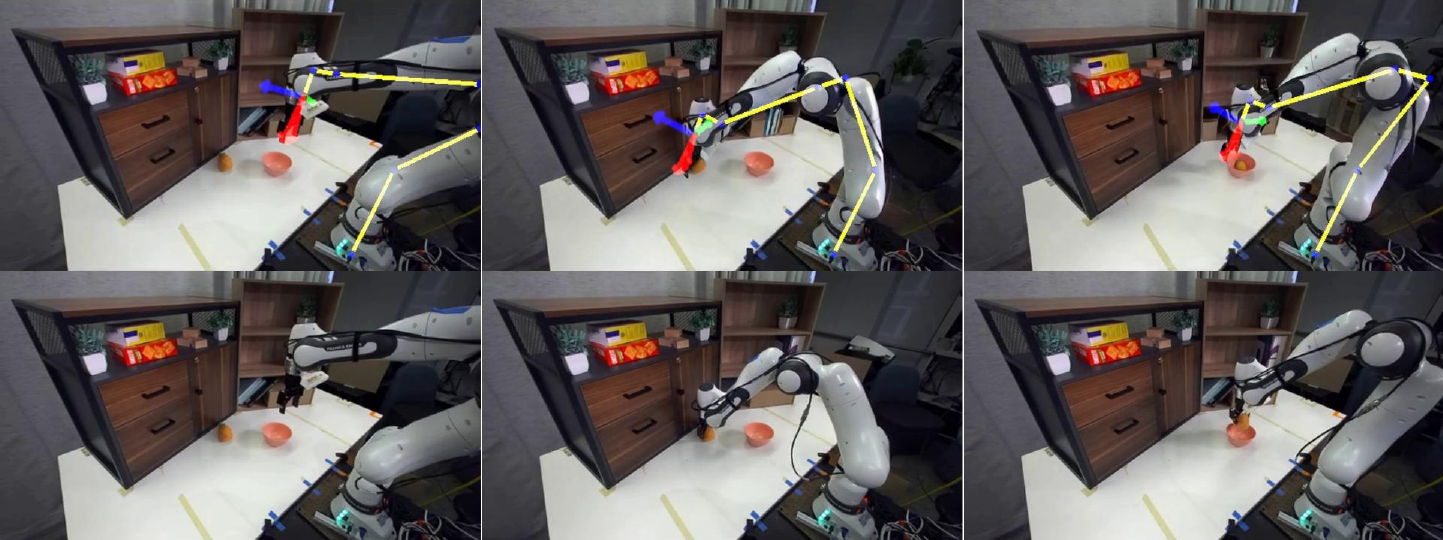}
  \end{minipage}\hfill
  \begin{minipage}[c]{0.44\linewidth}
    \centering
    \includegraphics[width=\linewidth]{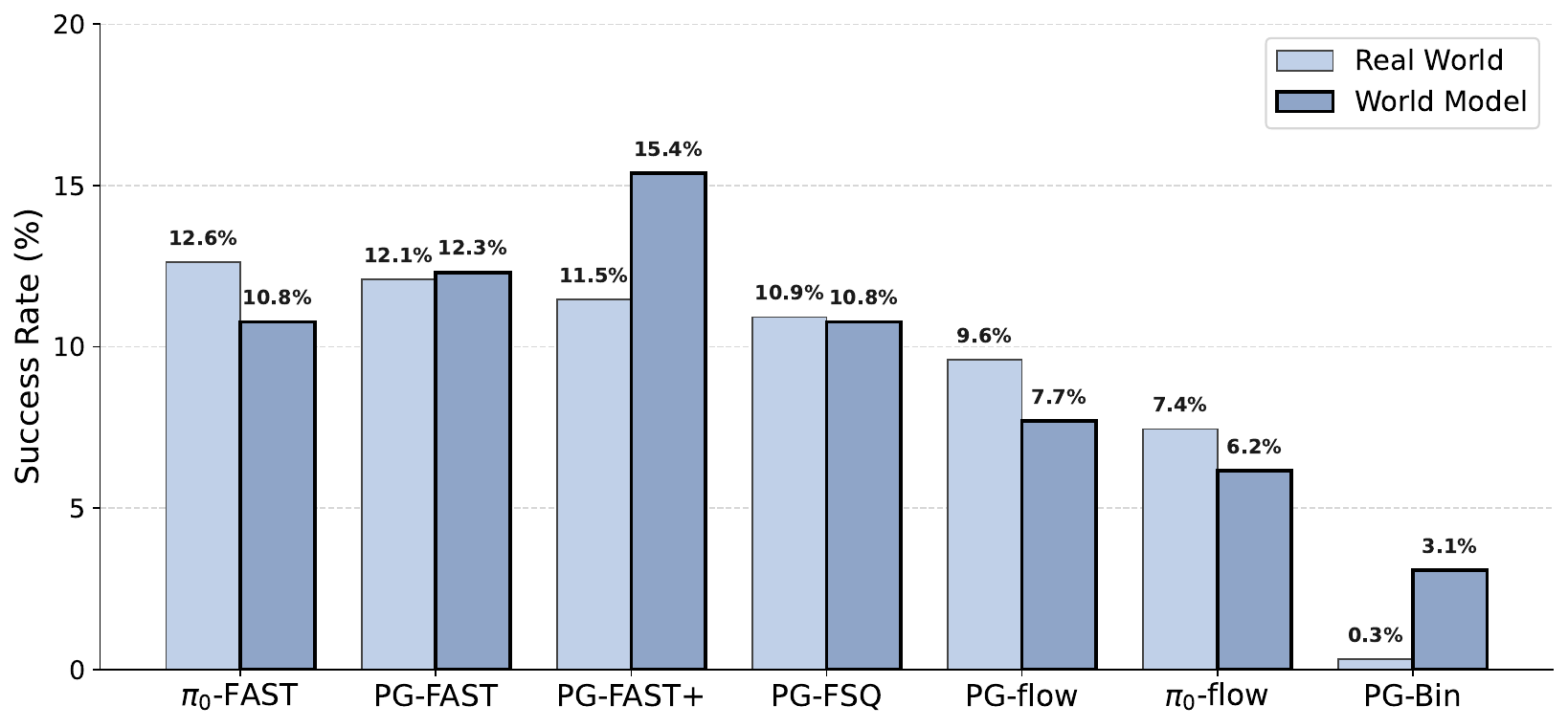}
  \end{minipage}
  \caption{\footnotesize \textbf{\methodname{} as a real-world policy-evaluation proxy on RoboArena~\citep{roboarena}.} \textbf{Left:} Comparison between a \methodname{} rollout (top) and the corresponding real-world rollout (bottom) for the $\pi_0$-FAST~\citep{pi0,fast} policy; three frames sampled uniformly over the episode. \textbf{Right:} Mean success rates on RoboArena across seven generalist policies: evaluating on our world model exhibits a strong correlation with real-world evaluation.}
  \label{fig:teaser}
\end{figure}

\begin{abstract}
  We present \methodname{}, a precise action-conditioned video world model that generalizes across different robot embodiments and enables robot policy evaluation.
Existing video world models face three main challenges for real-world robot evaluation: limited scenario diversity in current robot training datasets, imprecise action following, and poor generalization across embodiments for broad adoption.
We tackle these challenges from two perspectives.
At its core is a large-scale standardized data pipeline that curates, filters, and deduplicates broad robotics and egocentric human datasets, yielding a clean joint-training dataset that spans diverse tasks, scenarios, actions, and robot embodiments.
To condition the video model, we adopt 2D kinematic skeleton rendering as a unified conditioning representation that generalizes across different robot arms or even human hands.
We finetune the Cosmos-Predict2.5-2B  model on a single GH200 GPU. 
Our model achieves significant improvement on action following, appearance quality, and motion consistency, compared to existing baselines, which either have a much larger model size~\citep{kinema4d} or require more GPUs~\citep{genie-envisioner}. We further deploy \methodname{} to evaluate robot policies from RoboArena~\cite{roboarena}.
Extensive experiments demonstrate the significant correlation between our virtual policy evaluation in \methodname{} and real-world evaluation, paving the way for the future where robot policies can be purely evaluated in virtual generated worlds.

\end{abstract}

\keywords{World models, Cross-embodiment robot learning, Policy evaluation}

\section{Introduction}
\label{sec:intro}
Recent progress in large-scale video diffusion models has significantly advanced the development of world models, which emerge as a key component in building generalist robots~\citep{cosmos-predict2p5,unipi,unisim,gr1,vpp,irasim,kinema4d,genie-envisioner}.
Conditioned on an action sequence from robots, world models are positioned to predict the future state, reasoning the consequences of the action. Such a dynamic forecasting capability not only enables the policy evaluation and reinforcement learning with the generated virtual environment, but also is becoming the cornerstone for the pretraining in robot action planning~\citep{dreamzero,dreamgen}.

However, building a generalizable action-conditioned world model to faithfully evaluate robot policies faces three interconnected challenges: \textbf{(I)} The generated video needs to precisely follow the action conditions to indicate when the action happens (frame-level) and where it happens (pixel-level) to provide meaningful signals in the downstream policy evaluation. 
\textbf{(I)} The model must be able to generalize across diverse scenarios spanning different tasks, environments, and action sequences to ensure comprehensive policy evaluation.
\textbf{(III)} Instead of focusing on a particular robot embodiment, the world model should be able to generalize across different embodiments for broader adoption.

Existing action-conditioned world models fall short on these requirements. We categorize prior works according to their approach to action conditioning: \emph{Latent-action} methods compress robot state and actions into a learned embedding for video model conditioning ~\citep{irasim,enerverseac,ctrlworld,adaworld,gr2,dreamdojo}.
While being able to handle multiple embodiments during training, the action following in these models is often imprecise, as the model must infer spatial-temporal motion from a compressed latent embedding.
On the other side, \emph{explicit-conditioning} methods render the action into a video that is aligned with the RGB frames, such as pointmaps~\citep{kinema4d,track2act}, 3D occupancy~\citep{orv}, 2D gripper state~\citep{enerverseac,genie-envisioner}, or 2D kinematic skeletons~\citep{vap}. This design trades off expressiveness and generalization. 
Dense pointmaps~\citep{kinema4d} can be accurate for in-domain robots but may overfit to appearance and degrade out-of-distribution evaluation. Conversely, end-effector or gripper-only renders~\citep{enerverseac, genie-envisioner} are more robust but missing whole-arm motion. 
Close to our work, VAP~\cite{vap} uses the skeleton rendering to balance between the two. Yet, VAP is constrained by its relatively limited data scale, which only has 104k robot clips across two embodiments and 200k human clips, without filtering.

In this work, we present \methodname{}, a step towards building a general-purpose action-conditioned world model for robotics with precise action following and cross-embodiment generalization. First, we build a large-scale standardized data pipeline that curates, filters, and annotates publicly available robotics datasets.
We specifically emphasize diversity during data processing and collect datasets with 4 different robot embodiments, along with egocentric human object interaction, which provides substantial visual, motion, and scene diversity for generalization. Second, following VAP~\citep{vap}, we leverage the skeleton rendering as a generalizable conditioning schema for different embodiments of robots.  Skeleton rendering offers two benefits: \textbf{(I)} It only depends on the kinematic chain, and changing the embodiment only updates the kinematic specification. A single representation can therefore capture different robots, humans, or any mixture of them. \textbf{(II)} As the skeleton rendering has no textures, the model must explicitly capture the relationship between the kinematics and the actual robot movement, mitigating the overfitting to specific robot textures.

We finetune the Cosmos-Predict-2.5-2B video model using our curated high-quality and diverse dataset, along with the effective skeleton conditioning schema. We train the model on a \textbf{single} GH200 GPU. Compared to existing baselines, which either have a much larger model size (e.g., 14 billion parameters~\citep{kinema4d}) or require much larger GPU resources~\citep{genie-envisioner}, our model consistently outperforms and achieves significant improvement on action following, appearance quality, and motion consistency. 
We further deploy our model to evaluate the success rate of robotics policy and compare it against the success rate that is obtained from real-world deployment from RobotArena. Extensive experiments demonstrate the significant correlation between our virtual policy evaluation with \methodname{} and real-world evaluation, paving the way for reducing the robot policy evaluation cost and speeding up policy development iterations.
We release code, data, and trained checkpoints; see our \href{https://wuzy2115.github.io/oscar-project-page/}{project page} for more details.

\section{Related Work}
\label{sec:related}

\paragraph{Video World Models for Robot Manipulation.}
Recent video world models for robot manipulation often start from generic image-/text-to-video generators (e.g., Cosmos~\citep{cosmos-predict2p5}, Wan~\citep{wan21}) and extend them with additional conditioning signals. Broadly, these conditioning signals fall into three categories. UniSim~\citep{unisim} and TesserAct~\citep{tessaract} finetune on robotics scenes and use text to guide the action generation. \textit{Latent action} injects a compact embedding of the robot state/action (e.g., end-effector pose or a learned latent action sequence) into the generator, providing implicit control over the rollout; representative examples include AdaWorld~\citep{adaworld} and DreamDojo~\citep{dreamdojo}, IRASim~\citep{irasim}, DreamZero~\citep{dreamzero}, and Ctrl-World~\citep{ctrlworld}. \textit{Explicit conditions} provide spatially aligned rendered inputs that guide the video at the pixel level, including pointmap renderings (Kinema4D~\citep{kinema4d}), occupancy renderings (ORV~\citep{orv}), gripper state renderings (EnerVerse-AC~\citep{enerverseac}, Genie-Envisioner~\citep{genie-envisioner}), and 2D skeleton projections of the manipulator (VAP~\citep{vap}). We adopt the skeleton-based interface.


\paragraph{Video World Models for Policy Evaluation.} SIMPLER~\citep{simpler} introduced simulation as a proxy for real-robot policy evaluation and ranked manipulation policies in a hand-crafted real-to-sim environment using the MMRV and Pearson-correlation protocol.
IRASim~\citep{irasim} extended this protocol to trajectory-conditioned video world models and reported strong correlates with real success rate on RT-1 demonstrations. Later video world models split into two evaluation modes: WorldEval~\citep{worldeval} and EnerVerse-AC~\citep{enerverseac} score policies open-loop from pre-recorded actions, while WorldGym~\citep{worldgym}, Ctrl-World~\citep{ctrlworld}, GE-Sim~\citep{genie-envisioner}, and Scalable Policy Evaluation~\citep{scalablepolicyeval} run the policy in closed loop, and the success rate is evaluated by VLMs or humans. EWMBench~\citep{ewmbench} supplies a complementary metric suite. Following this protocol, we evaluate \methodname{} on the public RoboArena~\citep{roboarena} leaderboard: using off-the-shelf generalist DROID policies as in WorldGym, we report Pearson correlation and MMRV between \methodname{} rollouts and the official BT/Elo ranking.

\section{Method}
\label{sec:method}

\subsection{Preliminaries}
\label{sec:prelim}
We build on Cosmos-Predict2.5~\citep{cosmos-predict2p5}, a 2B video Diffusion Transformer (DiT) trained with a rectified-flow objective. A WAN~2.1 VAE~\citep{wan21} first encodes an $H{\times}W$ video $V_{1:T}$ into a spatio-temporal latent $z \in \mathbb{R}^{T'\times H'\times W'\times d}$.
The DiT flattens this latent into patch tokens and denoises them. The model is trained to predict a velocity field $v_\theta$ between the noise $\epsilon \sim \mathcal{N}(0, I)$ and the target latent $z_0$:
\begin{equation}
    \mathcal{L}_{\mathrm{RF}} \;=\; \mathbb{E}_{t,\,z_0,\,\epsilon}\,\big\| v_\theta(z_t,\, t,\, c) - (\epsilon - z_0) \big\|_2^2,
    \qquad z_t = (1-t)\,z_0 + t\,\epsilon,
    \label{eq:rf}
\end{equation}
where $c$ is the condition and is typically a text prompt, or an image. When conditioning on an image $I_0$, Cosmos keeps the first frame as the given image. Specifically, at every denoising step, the latent at the first temporal position is overwritten with the clean VAE encoding of $I_0$, and the model only denoises the future frames. We refer the readers to the original paper for further details.

\begin{figure*}[t]
\centering
\includegraphics[width=\linewidth]{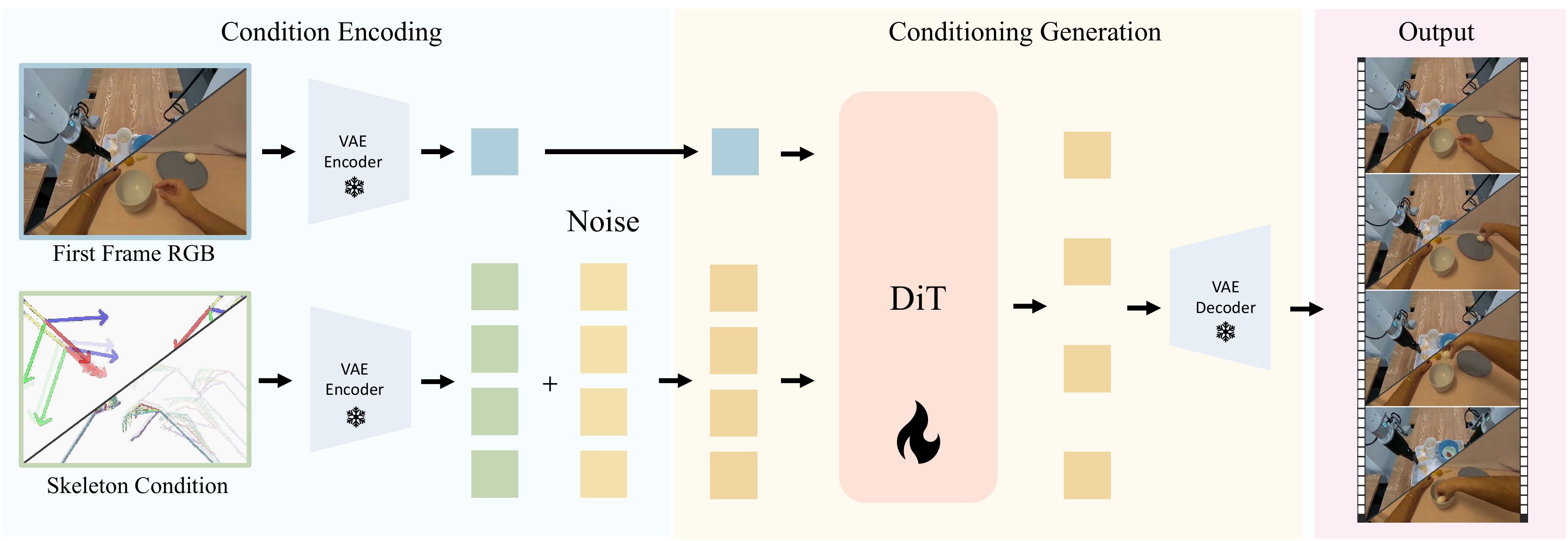}
\caption{\footnotesize \textbf{Method overview.} \methodname{} consists of three components: \textbf{(1) Condition encoding} encodes the first frame $I_0$ and rendered skeleton $S_{1:T}$ into latents using VAE; \textbf{(2) Conditioning injection} combines the skeleton latent with the noisy video latent; and \textbf{(3) Video generation}, where a DiT denoises the tokens and a VAE decoder decodes the final video.}
\label{fig:method}
\end{figure*}

\definecolor{cDROID}{HTML}{52AADC}
\definecolor{cRH20T5}{HTML}{EC6E66}
\definecolor{cRH20T7}{HTML}{D75B4E}
\definecolor{cIntern}{HTML}{2D8875}
\definecolor{cAgiBot}{HTML}{7C7979}
\definecolor{cAIROA}{HTML}{C89736}
\definecolor{cEgoDex}{HTML}{963B79}
\definecolor{cEPIC}{HTML}{BD7795}
\definecolor{cEgo4DCook}{HTML}{5C8857}
\definecolor{cEgo4DOther}{HTML}{886F3F}

\begin{figure*}[t]
\centering
\setlength{\tabcolsep}{2pt}
\renewcommand{\arraystretch}{1.0}
\begin{tabular}{@{}cccc@{}}
\includegraphics[width=0.245\linewidth]{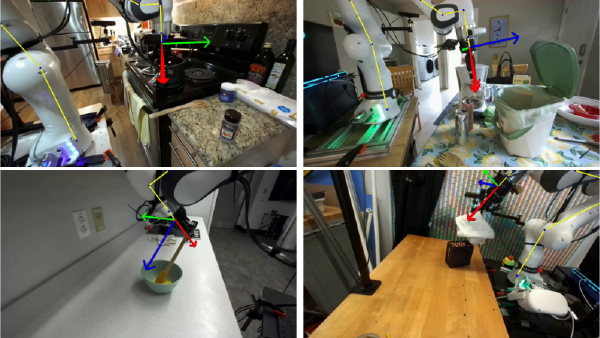} &
\includegraphics[width=0.245\linewidth]{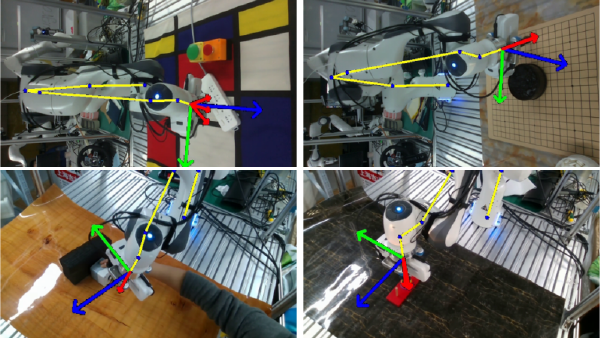} &
\includegraphics[width=0.245\linewidth]{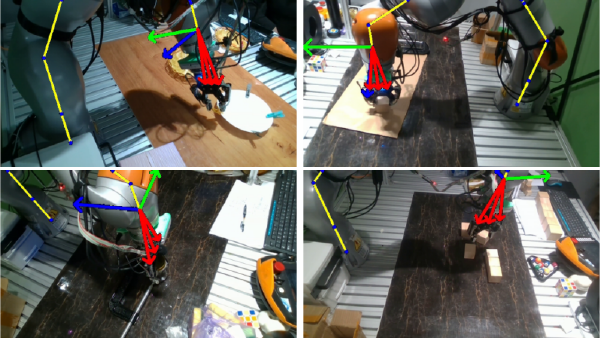} &
\includegraphics[width=0.245\linewidth]{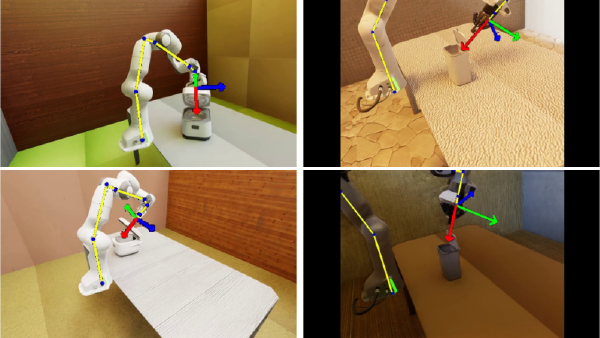} \\
\textbf{\textcolor{cDROID}{DROID}} &
\textbf{\textcolor{cRH20T5}{RH20T-cfg5}} &
\textbf{\textcolor{cRH20T7}{RH20T-cfg7}} &
\textbf{\textcolor{cIntern}{InternData}} \\[2pt]
\includegraphics[width=0.245\linewidth]{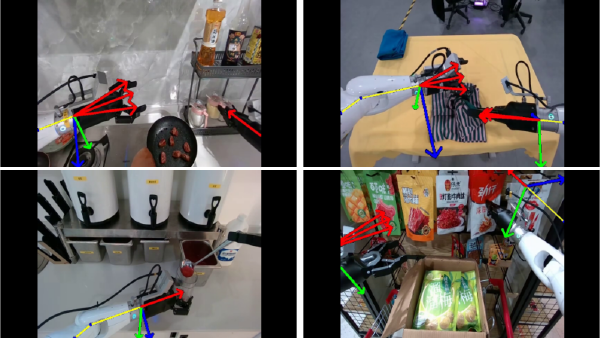} &
\includegraphics[width=0.245\linewidth]{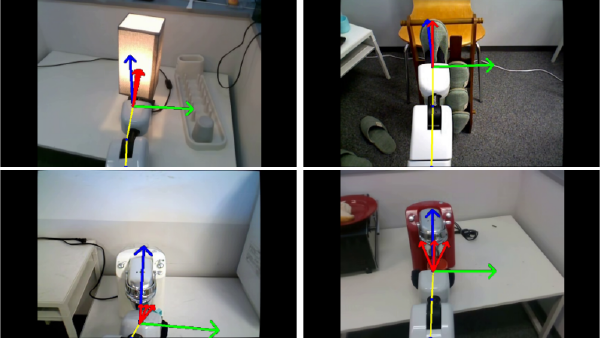} &
\includegraphics[width=0.245\linewidth]{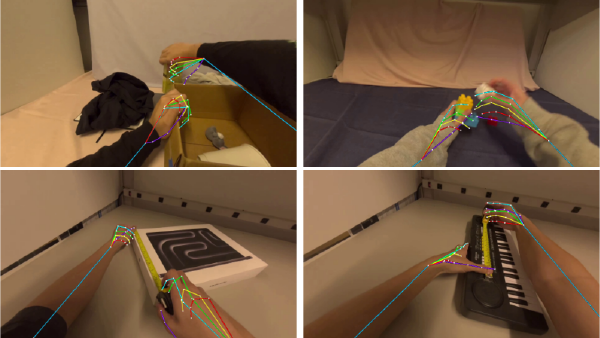} &
\includegraphics[width=0.245\linewidth]{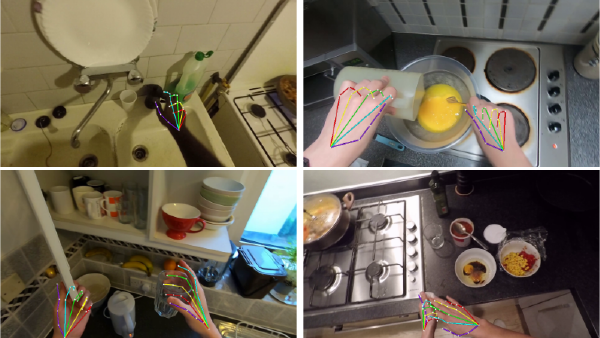} \\
\textbf{\textcolor{cAgiBot}{AgiBot G1}} &
\textbf{\textcolor{cAIROA}{AIROA-MoMa}} &
\textbf{\textcolor{cEgoDex}{EgoDex}} &
\textbf{\textcolor{cEPIC}{EPIC-Kitchens}} \\
\end{tabular}
\caption{\footnotesize Skeleton overlays at video frames for the eight training sources. Each block shows four episodes from one source. \textbf{Top row}: DROID, RH20T-cfg5, RH20T-cfg7, InternData (four robot recordings). \textbf{Bottom row}: AgiBot G1, AIROA-MoMa, EgoDex, EPIC-Kitchens (humanoid and two human MANO sources).}
\label{fig:skeleton_grid}
\end{figure*}

\subsection{Skeleton Rendering as a Unified Conditioning}
\label{sec:control}
Choosing the right action representation for video conditioning is the cornerstone for the action-conditioned world model. Many prior approaches face a practical trade-off between \emph{generalization} and \emph{precision}. Specifically, latent-action representations~\citep{irasim, adaworld, ctrlworld, dreamdojo, dreamzero} can represent multiple embodiments of robots, but the implicit action signal typically generates results that can not precisely follow the given action, especially when the target motion differs from the training distribution. On the other hand, more detailed renderings of robot geometry (e.g. mesh or pointmap)~\citep{kinema4d, track2act} can improve precision in action following. Yet, they often entangle embodiment-specific appearance with motion, and may hurt cross-robot generalization.

In our paper, we choose skeleton renderings as the action representation for robotics, as they balance well between \emph{generalization} and \emph{precision}. By only rasterizing the projected kinematic tree (including a visual indicator of gripper state), the skeleton renderings provide explicit guidance on robot action, while remaining largely invariant to arms' textures and materials. We intentionally avoid adding fine-grained surface textures in the action condition, as the first RGB frame $I_0$ typically anchors the robot's appearance and the scene.
We detail the skeleton rendering pipeline below.

\paragraph{Skeleton Rendering.}
Figure~\ref{fig:skeleton_grid} shows an visualization of our skeleton rendering on eight training datasets we collected (Sec.~\ref{sec:data}). Let $M$ denote the URDF model with kinematic tree $(\mathcal{V}(M), \mathcal{E}(M))$, where $\mathcal{V}(M)$ collects its $K$ links and $\mathcal{E}(M)$ the parent--child edges. Given the joint configuration $q_t$ at time $t$, forward kinematics yields one SE(3) pose per link,
$\big\{T_{k,t}\big\}_{k=1}^{K} \;=\; \mathrm{FK}(q_t,\, M).
$
Picking a canonical point $o_k$ per link (the link origin in its own frame), its pixel projection under camera intrinsics $K_\mathrm{cam}$ and extrinsic $T^{\mathrm{cam}}_{\mathrm{world}}$ is:
    $\big(u_{k,t},\, v_{k,t}\big) \;=\; \pi\!\Big(K_\mathrm{cam},\; T^{\mathrm{cam}}_{\mathrm{world}}\, T_{k,t}\, o_k\Big)$,
where $\pi$ is the standard perspective projection. 
We then rasterise the projected kinematic tree onto a black canvas:
    $S_t \;=\; \mathrm{Rasterise}\!\Big(\big\{(u_{k,t},\, v_{k,t})\big\}_{k=1}^{K},\; \mathcal{E}(M)\Big),
    \label{eq:raster}$
$\mathrm{Rasterise}(\cdot)$ operates entirely in pixel space: it draws a line segment between the projected endpoints of each edge in $\mathcal{E}(M)$ and a small filled circle at every projected vertex $(u_{k,t}, v_{k,t})$. All other pixels remain black. The result carries only 2D visualization of the kinematic chain, and is the cheapest geometry information that indicates the action.

\paragraph{Conditioning Injection.}
We feed the skeleton rendering $S_{1:T}$ to the DiT as a second RGB video stream aligned frame-by-frame with the target (Figure~\ref{fig:method}). Specifically, we pass it through the same WAN~2.1 VAE and produce a skeleton latent $z^s$ with the same shape as the target video latent $z^v_t$. The two latents are then embedded into the DiT hidden dimension with a patch embedder $\mathrm{PE}_v$ and $\mathrm{PE}_s$, respectively. The resulting token tensors are summed together to feed into the DiT for denoising.

\paragraph{Extension to Human Hands.}
Since $S_{1:T}$ encodes only 2D joint projections, the same conditioning representation can be used for both robot arms and human hands (Figure~\ref{fig:method}). In practice, we keep most rendering process unchanged and only swap the kinematic triple $(M, q_t, o_k)$ for human hands. Specifically, for a MANO hand model with topology $M^{\mathrm{MANO}}$~\citep{mano}, per-frame pose parameters $q_t^{\mathrm{MANO}}$, and canonical joint points $o_k^{\mathrm{MANO}}$,
    $S^{\mathrm{human}}_{t} \;=\; \mathrm{Rasterise}\!\Big(\big\{\pi\!\big(K_\mathrm{cam},\, T^{\mathrm{cam}}_{\mathrm{world}}\, T^{\mathrm{MANO}}_{k,t}\, o_k^{\mathrm{MANO}}\big)\big\}_{k},\;\mathcal{E}(M^{\mathrm{MANO}})\Big),
    \label{eq:human-skel}$
where $\{T^{\mathrm{MANO}}_{k,t}\}_k = \mathrm{FK}(q_t^{\mathrm{MANO}},\, M^{\mathrm{MANO}})$. Although a five-finger hand has more DoFs than a two-jaw gripper, both are rendered into the same 2D line drawings; the skeleton therefore constrains the coarse motion, while the pretrained video prior fills in plausible fine-grained visual details. This allows us to incorporate broader human egocentric demonstrations as an additional dataset during training, significantly increasing the diversity in the scenarios, tasks, and action distributions. 


\section{Data Pipeline}
\label{sec:data}


Training action-conditioned world models requires a diverse, high-quality, and large-scale dataset that can cover broad scenarios and tasks in the real world. However, most existing robotics datasets have narrow distributions on the environments, objects, tasks, or robot embodiments~\cite{agibotworld, rh20t, interndataa1, droid, airoamoma}. A prominent example is AgiBot~\cite{agibotworld}. Although it has 1 million clips, most of the clips cover almost the same environments or tasks. We, therefore, introduce a four-stage data pipeline to construct a diverse and large-scale robotic video dataset. After processing, we filtered out 180{,}657 episodes out of 2{,}165{,}359 source videos; Table~\ref{tab:data_sources} reports statistics and Figure~\ref{fig:skeleton_grid} shows representative samples.

\subsection{Data Curation}
\label{sec:data_sources}
\begin{wraptable}[12]{r}{0.48\textwidth}
\vspace{-9mm}
\caption{\footnotesize\raggedright Data statistics (episodes). Public: official dataset scale; Filtered: after our filters.}
\centering
\label{tab:data_sources}
\scriptsize
\setlength{\tabcolsep}{4pt}
\begin{tabular*}{\linewidth}{@{\extracolsep{\fill}}llrr}
\toprule
& & \multicolumn{1}{c}{Public} & \multicolumn{1}{c}{Filtered} \\
Source & Embodiments & Episodes & Episodes \\
\midrule
RH20T (cfg5)\textsuperscript{\hyperref[note:rh20t]{2}}        & Franka Panda             & 2{,}241        & 1{,}261 \\
RH20T (cfg7)\textsuperscript{\hyperref[note:rh20t]{2}}        & KUKA iiwa                & ---            & --- \\
InternData-A1\textsuperscript{\hyperref[note:synthetic]{1}}       & Franka Panda             & 630{,}000      & 2{,}233 \\
DROID                     & Franka Panda             & 76{,}000       & 21{,}904 \\
AgiBot-Beta               & AgiBot G1                & 1{,}003{,}672  & 65{,}720 \\
AIROA-MoMa                & Toyota HSR               & 25{,}469       & 3{,}712 \\
\multicolumn{2}{r}{\textit{Robot subtotal}} & \textbf{1{,}737{,}382} & \textbf{94{,}830} \\
\midrule
EgoDex              & human hand               & 338{,}000      & 78{,}273 \\
EPIC-Kitchens       & human hand               & 89{,}977       & 7{,}554 \\
\multicolumn{2}{r}{\textit{Human subtotal}} & \textbf{427{,}977} & \textbf{85{,}827} \\
\midrule
\multicolumn{2}{r}{\textbf{Total}} & \textbf{2{,}165{,}359} & \textbf{180{,}657} \\
\bottomrule
\end{tabular*}
\vspace{-\intextsep}
\end{wraptable}
We curate from both robotics and human videos as our data source to ensure broad scene coverage. Specifically, we select five robot datasets and two egocentric human videos. Details are provided in Table~\ref{tab:data_sources}.
For robotics data, we curate from
RH20T-cfg5~\citep{rh20t}  (Franka Panda), InternData-A1~\citep{interndataa1}, DROID~\citep{droid},
RH20T-cfg7 (KUKA iiwa),
AgiBot~\citep{agibotworld} (AgiBot G1),
and AIROA-MoMa~\citep{airoamoma} (Toyota HSR). In total, the public \emph{released} robot sources amount to about 10.9k hours across about 1.74M episodes.
For the human data, we curate from EgoDex~\citep{egodex} and EPIC-Kitchens~\citep{vitra,damen2020epic}, which amount to about 929 hours across about 428k episodes.
Further details on per-source collection procedures and dataset-specific considerations are provided in the Appendix~\ref{app:dataset_details}.

\subsection{Data Filtering}
\label{sec:data_filtering}

Robotics video collected at scale is inevitably heterogeneous and noisy: episodes may be too short, dominated by camera motion, nearly static, partially out-of-view, or corrupted by sensor artefacts. Such clips can bias a video diffusion model toward degenerate solutions (e.g., learning freeze-frame).
We apply the following four mechanism-specific filters to both human and robot data.

\begin{enumerate}
\item \textbf{Length.} Each clip must contain at least $70$ video frames to ensure enough rollout.

\item \textbf{Static Camera.} We mainly focus on the robot action in our paper and defer the camera motion for future work. Thus, we only keep episodes with static cameras. We achieve this by filtering out episodes whose camera movements are larger than a threshold. 

\item \textbf{Meaningful Action.} Each episode must contain a non-trivial manipulator action sequence. This helps us remove purely static action episodes.


\item \textbf{Visible Skeleton.} The robot skeleton should remain visible within the video frames. We filter out episodes whose visible skeleton percentages are lower than a threshold.

\end{enumerate}

\subsection{Semantic Deduplication}
\label{sec:dedup_pipeline}

After the quality filters, we observe substantial redundancy in both robotics and egocentric human data: they often repeat many highly similar tasks in the same physical scene. This inflates the raw clip count while adding little scene diversity. Our goal is to expose the video model to diverse environments for generalization. We deduplicate along two complementary axes: \textbf{visual redundancy}, which captures repeated scenes and backgrounds, and \textbf{trajectory redundancy}, which captures repeated robot or hand motion. We do \emph{not} treat clips as duplicates if they share a scene but follow substantially different trajectories. We perform semantic deduplication with a two-stage pipeline that first clusters visually similar candidates and then verifies them using trajectory similarity.

\paragraph{Stage 1: Visual Clustering.}
For each episode, we compute a SigLIP~\citep{siglip} image embedding from five uniformly sampled frames. We compare the pair-wise visual similarity by SigLIP cosine similarity, flagging every pair above $0.95$ for the following trajectory verification.

\paragraph{Stage 2: Trajectory Verification.}
We also extract a $64$-step resampled manipulator trajectory 
from each episode. For each pair with high visual similarity, we compute the per-step root-mean-square (RMS) distance between action trajectories. We mark a pair as a duplicate only if its RMS falls below an adaptive threshold to avoid filtering episodes with similar background but diverse motion. 


\subsection{Data Captioning}
\label{sec:data_captioning}

We caption all retained episodes with Qwen3-VL-30B-A3B-Instruct~\citep{qwen3vl}.  For each episode, we sample input frames at $15$\,fps. Since episodes from DROID are typically $5\times$ longer than other sources, we drop their sampling rate to $1$--$2$\,fps for efficiency.

\begin{table}[t]
\centering
\caption{\footnotesize Quantitative comparison with baselines. We report the metrics averaged over all four embodiments. 
Per-embodiment results are listed in Appendix~\ref{app:per_emb}. 
\textbf{Best} results are in \textbf{bold} and \underline{second-best} results are underlined.
}
\label{tab:main_quant}
\small
\setlength{\tabcolsep}{6pt}
\begin{tabular}{lcccccccc}
\toprule
Method & PSNR$\uparrow$ & SSIM$\uparrow$ & LPIPS$\downarrow$ & tLPIPS$\downarrow$ & FVD$\downarrow$ & FID$\downarrow$ & $\text{L2}_{\text{latent}}$$\downarrow$ & FPS$\uparrow$ \\
\midrule
Cosmos-Predict2.5~\citep{cosmos-predict2p5}    & 14.78 & 0.563 & 0.370 & 0.022 & 18.01 & 47.59 & 0.435 & 0.292 \\
TesserAct~\citep{tessaract}                    & 16.26 & 0.730 & 0.277 & 0.055 & 24.50 & 51.90 & 0.364 & 0.343 \\
\addlinespace[2pt]
IRASim$^{1}$~\citep{irasim}                    & 6.48 & 0.088 & 0.909 & 0.606 & 411.42 & 394.10 & 2.453 & \textbf{2.330} \\
Ctrl-World~\citep{ctrlworld}                   & 19.06 & 0.705 & 0.321 & 0.042 & 28.90 & 53.33 & 0.292 & 1.631 \\
\addlinespace[2pt]
EnerVerse-AC~\citep{enerverseac}               & 20.47 & 0.746 & 0.223 & 0.021 & 33.70 & 38.23 & 0.197 & 1.900 \\
Genie Envisioner~\citep{genie-envisioner}      & \underline{23.29} & \underline{0.838} & \underline{0.140} & \textbf{0.007} & \underline{15.37} & \underline{22.92} & \underline{0.129} & 1.382 \\
Kinema4D~\citep{kinema4d}                      & 17.68 & 0.741 & 0.198 & 0.021 & 17.07 & 37.16 & 0.233 & 0.089 \\
\midrule
\methodname{} (\textbf{Ours})                  & \textbf{24.24} & \textbf{0.846} & \textbf{0.094} & \underline{0.015} & \textbf{7.08} & \textbf{15.07} & \textbf{0.096} & \underline{2.214} \\ 
\bottomrule
\end{tabular}
\end{table}
\section{Experiments}
\label{sec:exp}
\subsection{Experimental Settings}
\label{sec:exp_setup}
\label{sec:training}
We finetune Cosmos-Predict2.5-2B~\citep{cosmos-predict2p5} in two stages. Stage 1 trains $15$k iterations on the four robot embodiments. Stage 2 continues on the full robot+human mixture. Further training details are provided in Appendix~\ref{app:training_setup}.
We evaluate on a self-curated benchmark of $200$ robot manipulation clips drawn from six datasets across four embodiments: Franka Panda, KUKA iiwa, AgiBot G1, and Toyota HSR. Selection rules are in Appendix~\ref{app:eval_benchmark}.
We compare against seven baselines categorized by their conditioning approach: text-only (TesserAct~\citep{tessaract}, Cosmos-Predict2.5~\citep{cosmos-predict2p5}), latent-action (IRASim~\citep{irasim}, Ctrl-World~\citep{ctrlworld}, EnerVerse-AC~\citep{enerverseac}), and explicit-geometry (Genie Envisioner~\citep{genie-envisioner}, Kinema4D~\citep{kinema4d}). 
We measure the quality from four perspectives: reconstruction quality (PSNR~\citep{PSNR}, SSIM~\citep{SSIM}, LPIPS~\citep{lpips}), temporal coherence (tLPIPS), distribution fidelity (FVD~\citep{fvd}, FID~\citep{fid}, latent-$\ell_2$), and speed (FPS). All models are timed on the same NVIDIA GH200 GPU. For a fair comparison, we compute the metrics on the first $49$ frames, following the Kinema4D protocol.

\subsection{Comparison with Baselines}
\label{sec:exp_quant}

Table~\ref{tab:main_quant} reports the quantitative results, and Figure~\ref{fig:qual_compare} provides a qualitative comparison. Further results are provided in Appendix~\ref{app:per_emb_qual}.
Overall, \methodname{} ranks best or second-best on most metrics and outperforms the $14$B-parameter Kinema4D (\methodname{} is only 2B). We make three observations. (i) \emph{Text-only} control~\citep{tessaract, cosmos-predict2p5, unisim} is weakest because language can not precisely describe the action sequence. (ii) \emph{Latent-action} guidance~\citep{irasim, ctrlworld} is limited by the training embodiments: a fixed kinematic layout becomes out-of-distribution when transferred between bimanual and single-arm setups. (iii) \emph{Explicit, spatially aligned} guidance~\citep{vap, enerverseac, genie-envisioner, kinema4d} 
performs best, with dense point-maps overfitting in-distribution and our skeleton condition giving the best accuracy/generalisation trade-off.

\begin{table}[t]
\centering
\caption{\footnotesize Ablations on conditioning representation and data composition. Top block (gray): we train the model with the same robot-only dataset and vary different conditioning representations. Bottom block: we use the same skeleton conditioning and incorporate human data at different training stages. Bold rows ($^\star$) mark the configurations adopted in the final model. ``+Human'' refers to the mixed training with human data.}
\label{tab:abl}
\small
\setlength{\tabcolsep}{5pt}
\begin{tabular}{llccccccc}
\toprule
Aspect & Variant & PSNR$\uparrow$ & SSIM$\uparrow$ & LPIPS$\downarrow$ & tLPIPS$\downarrow$ & FVD$\downarrow$ & FID$\downarrow$ & $\text{L2}_{\text{latent}}\downarrow$ \\
\midrule
\rowcolor{gray!15} \multirow[t]{3}{*}{\cellcolor{gray!15}Condition}
 & Latent action        & 19.22 & 0.784 & 0.170 & 0.018 & 12.03 & 26.11 & 0.205 \\
\rowcolor{gray!15}
 & Mesh rendering       & 23.11 & 0.831 & 0.106 & 0.013 & 7.89 & 16.38 & 0.109 \\
\rowcolor{gray!15}
 & \textbf{Skeleton (canonical)}$^\star$ & \textbf{23.48} & \textbf{0.832} & \textbf{0.106} & \textbf{0.015} & \textbf{7.69} & \textbf{16.37} & \textbf{0.117} \\
\midrule
\rowcolor{gray!3} \multirow[t]{2}{*}{\cellcolor{gray!3}Data mix}
 & +Human, from beginning   & 23.87 & 0.842 & 0.097 & 0.014 & 7.65 & 15.72 & 0.100 \\
\rowcolor{gray!3}
 & \textbf{+Human, warm-start}$^\star$ & \textbf{24.24} & \textbf{0.846} & \textbf{0.094} & \textbf{0.015} & \textbf{7.08} & \textbf{15.07} & \textbf{0.096} \\
\bottomrule
\end{tabular}
\end{table}

\begin{figure}[t]
\centering
\vspace{-6pt}
\begin{tikzpicture}
  \node[anchor=south west,inner sep=0pt] (qa) {\includegraphics[width=0.95\linewidth]{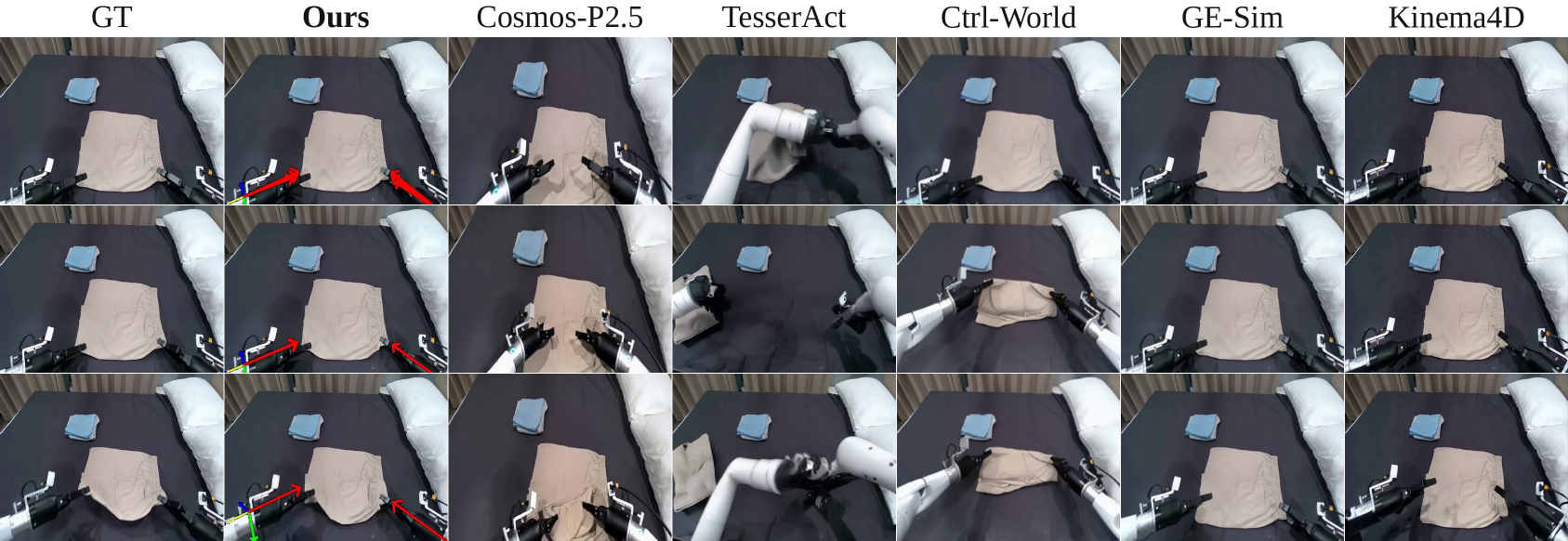}};
  \coordinate (qaft) at ($(qa.north west)!0.068!(qa.south west)$); 
  \draw[-{Stealth[length=5pt,width=4pt]},line width=1.1pt,cAgiBot]
    ([xshift=-4pt]qaft) -- ([xshift=-4pt]qa.south west);
  \node[rotate=90,anchor=center,inner sep=0pt,minimum height=14pt] at ([xshift=-13pt]$(qaft)!0.5!(qa.south west)$)
    {\textbf{\textcolor{cAgiBot}{AgiBot G1}}};
\end{tikzpicture}\\[-1pt]
\begin{tikzpicture}
  \node[anchor=south west,inner sep=0pt] (qd) {\includegraphics[width=0.95\linewidth,trim=0 0 0 18.96bp,clip]{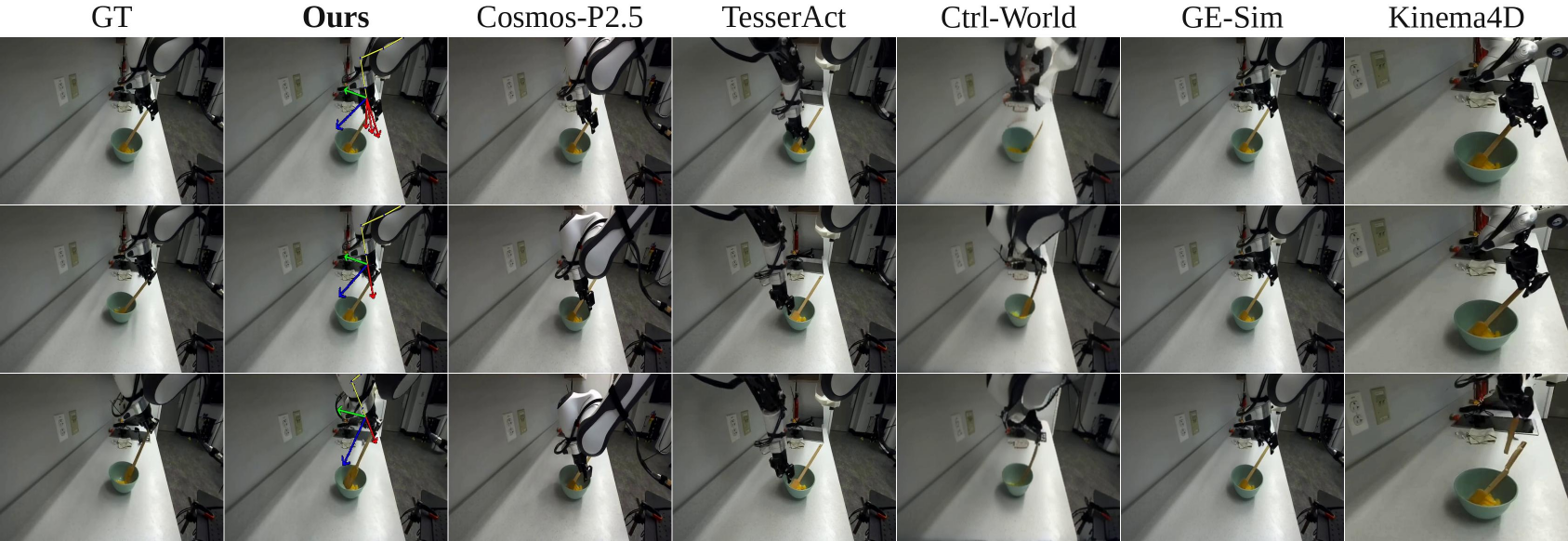}};
  \draw[-{Stealth[length=5pt,width=4pt]},line width=1.1pt,cDROID]
    ([xshift=-4pt]qd.north west) -- ([xshift=-4pt]qd.south west);
  \node[rotate=90,anchor=center,inner sep=0pt,minimum height=14pt] at ([xshift=-13pt]qd.west)
    {\textbf{\textcolor{cDROID}{DROID}}};
\end{tikzpicture}
\caption{\footnotesize Qualitative comparison of action-conditioned video generation on two embodiments. Compared with five baselines, our method achieved much better visual quality with precise action following.}
\label{fig:qual_compare}
\vspace{-10pt}
\end{figure}

\subsection{Ablation Studies}
\label{sec:exp_ablation}
We ablate two factors central to \methodname{}. First, alternative \emph{conditioning representations} to compare with latent-action and mesh renderings. Second, \emph{data strategy} to analyze the effect of human videos. 

\paragraph{Conditioning Representation.}
We finetune the same Cosmos-2.5 model using purely robotics datasets, and only vary the conditioning representation. Quantitative results are in the top block of Table~\ref{tab:abl}, with qualitative results in Appendix~\ref{app:abl_qual}. Latent action fails to follow precise action cues. Mesh and skeleton are statistically indistinguishable across all seven metrics, but mesh depends on robot-specific URDF assets. We choose skeleton rendering as it allows us to incorporate human data.

\paragraph{Data Strategy.}
\label{sec:exp_ablation_data}

The bottom block of Table~\ref{tab:abl} provides quantitative results.
Adding human data into the training data mixture consistently improves the performance over robot-only training,  
indicating positive transfer from human to robots. Moreover, when continuing from the robot-only model, warm-starting accelerates model convergence. We thus choose this strategy as our final model.



\subsection{Policy Evaluation}
\label{sec:exp_policy_eval}
\begin{wraptable}[7]{r}{0.48\textwidth}
\vspace{-8mm}
\caption{\footnotesize Policy-evaluation using \methodname{} with three conditioning representations on the 65-session $\times$ 7-policies from RoboArena pool. Lower MMRV / $\mathrm{SISR}_{\Delta}$ and higher $\rho$ / $r$ are better.}
\centering
\label{tab:policy_eval}
\scriptsize
\setlength{\tabcolsep}{4pt}
\begin{tabular*}{\linewidth}{@{\extracolsep{\fill}}lcccc}
\toprule
Condition & MMRV $\downarrow$ & $\rho$ $\uparrow$ & $r$ $\uparrow$ & $\mathrm{SISR}_{\Delta}$ (pp) $\downarrow$ \\
\midrule
Latent action & 1.429 & $+0.643$ & $\mathbf{+0.867}$ & 1.98 \\
Mesh          & 0.714 & $+0.679$ & $+0.781$         & 3.04 \\
Skeleton      & $\mathbf{0.571}$ & $\mathbf{+0.750}$ & $+0.852$ & $\mathbf{1.73}$ \\
\bottomrule
\end{tabular*}
\vspace{-0.8cm}
\end{wraptable}
We deploy \methodname{} to evaluate real-robot policy from RoboArena leaderboard~\citep{roboarena}, which ranks seven open-sourced DROID generalist policies (\mbox{$\pi_0$-flow}, \mbox{$\pi_0$-FAST}, \mbox{PG-flow}, \mbox{PG-FSQ}, \mbox{PG-FAST}, \mbox{PG-FAST+}, \mbox{PG-Bin})~\citep{pi0, fast}.
We manually retain $65$ sessions for all 7 policies due to their decent camera calibration estimation.
For each session, we estimate camera intrinsics with MoGe-v2~\citep{mogev2} and cam-to-base extrinsics with CtRNet-X~\citep{ctrnetx}. 
For each episode, we autoregressively roll out \methodname{} from the recorded first frame and the given robot action. We then prompt GPT-5 to evaluate the success rate for each episode and compute its Pearson correlation ($r$~$\uparrow$), and the difference ($\mathrm{SISR}_{\Delta}\downarrow$) with the success rate in the real-world deployment. To further demonstrate the effectiveness of evaluating robotics policy in video models, we prompt GPT-5 to rank pair-wise policies in the same session by feeding two generated videos together, following RoboArena~\cite{roboarena} and WorldEval~\citep{worldeval}. This gives us $1\,365$ pairwise preferences in total, and we compute per-policy Bradley--Terry scores, and success rates against results from the RoboArena leaderboard. Following SIMPLER~\citep{simpler} and WorldGym~\citep{worldgym}, we report rank fidelity (MMRV~$\downarrow$, Spearman $\rho$~$\uparrow$). Quantitative results are provided in Table~\ref{tab:policy_eval}.
Evaluating robotics policy using our \methodname{} has a significant correlation with the real-world deployment, demonstrating the strong potential of using world models for robot policy evaluation.
We further compare three conditioning representations: latent-action, mesh renderings, and our skeleton renderings following \S\ref{sec:exp_ablation}. Our skeleton rendering provides the strongest correlation with the real-world deployment, demonstrating the effectiveness of our condition representation.
Further analysis and visualizations are in Appendix~\ref{app:policy_eval}.

\section{Conclusion}
\label{sec:conclusion}
We presented \methodname{}, an action-conditioned video world model with precise action following and cross-embodiment generalization for robot policy evaluation.
Two designs drive these properties.
First, a standardized data pipeline curates, filters, and deduplicates raw robotics and egocentric human video into a clean joint-training corpus that spans diverse tasks, scenes, actions, and four robot embodiments plus the human hand.
Second, 2D kinematic skeleton rendering serves as a single conditioning representation across robot arms and human hands: changing the embodiment only updates the kinematic specification, and the texture-free render keeps the model from binding motion to specific robot appearance.
We finetuned Cosmos-Predict2.5-2B on a single GH200 GPU, and our model outperforms baselines that use far more parameters~\citep{kinema4d} or GPUs~\citep{genie-envisioner} on action following, appearance quality, and motion consistency.
Deployed on RoboArena~\cite{roboarena}, virtual policy evaluation with \methodname{} correlates strongly with real-world evaluation, a step toward evaluating robot policies in generated worlds and cutting evaluation cost.



\paragraph{Limitations.}
Our current data scale is limited by the availability and quality of per-dataset camera calibration and kinematic annotations: errors in camera intrinsics/extrinsics directly degrade skeleton--RGB alignment, limiting the availability of raw video that can be reliably converted into usable training dataset. In addition, our model only uses a 2B-parameter backbone; scaling to larger backbones may further improve fidelity and generalization but requires more compute. 

\clearpage

{
\small
\bibliography{refs}

\begin{thebibliography}{50}
\providecommand{\natexlab}[1]{#1}
\providecommand{\url}[1]{\texttt{#1}}
\expandafter\ifx\csname urlstyle\endcsname\relax
  \providecommand{\doi}[1]{doi: #1}\else
  \providecommand{\doi}{doi: \begingroup \urlstyle{rm}\Url}\fi

\bibitem[Atreya et~al.(2025)Atreya, Pertsch, Lee, Kim, Jain, Kuramshin, Neary,
  Hu, Arora, Ellis, et~al.]{roboarena}
P.~Atreya, K.~Pertsch, T.~Lee, M.~J. Kim, A.~Jain, A.~Kuramshin, C.~Neary,
  E.~S. Hu, K.~Arora, K.~Ellis, et~al.
\newblock Roboarena: Distributed real-world evaluation of generalist robot
  policies.
\newblock In \emph{Conference on Robot Learning}, pages 336--364. PMLR, 2025.

\bibitem[Black et~al.(2024)Black, Brown, Driess, Esmail, Equi, Finn, Fusai,
  Groom, Hausman, Ichter, et~al.]{pi0}
K.~Black, N.~Brown, D.~Driess, A.~Esmail, M.~Equi, C.~Finn, N.~Fusai, L.~Groom,
  K.~Hausman, B.~Ichter, et~al.
\newblock {$\pi_0$}: A vision-language-action flow model for general robot
  control.
\newblock \emph{arXiv preprint arXiv:2410.24164}, 2024.

\bibitem[Pertsch et~al.(2025)Pertsch, Stachowicz, Ichter, Driess, Nair, Vuong,
  Mees, Finn, and Levine]{fast}
K.~Pertsch, K.~Stachowicz, B.~Ichter, D.~Driess, S.~Nair, Q.~Vuong, O.~Mees,
  C.~Finn, and S.~Levine.
\newblock Fast: Efficient action tokenization for vision-language-action
  models.
\newblock \emph{arXiv preprint arXiv:2501.09747}, 2025.

\bibitem[Xu et~al.(2026)Xu, Zhang, Liu, Chen, Han, and Liu]{kinema4d}
M.~Xu, T.~Zhang, T.~Liu, Z.~Chen, X.~Han, and Z.~Liu.
\newblock Kinema4{D}: Kinematic 4{D} world modeling for spatiotemporal embodied
  simulation.
\newblock \emph{arXiv preprint arXiv:2603.16669}, 2026.

\bibitem[Liao et~al.(2025)Liao, Zhou, Huang, Yang, Chen, Jiang, Hu, Cai, Liu,
  Luo, et~al.]{genie-envisioner}
Y.~Liao, P.~Zhou, S.~Huang, D.~Yang, S.~Chen, Y.~Jiang, Y.~Hu, J.~Cai, S.~Liu,
  J.~Luo, et~al.
\newblock Genie envisioner: A unified world foundation platform for robotic
  manipulation.
\newblock \emph{arXiv preprint arXiv:2508.05635}, 2025.

\bibitem[Ali et~al.(2025)Ali, Bai, Bala, Balaji, Blakeman, Cai, Cao, Cao, Cha,
  Chao, and other]{cosmos-predict2p5}
A.~Ali, J.~Bai, M.~Bala, Y.~Balaji, A.~Blakeman, T.~Cai, J.~Cao, T.~Cao,
  E.~Cha, Y.-W. Chao, and other.
\newblock World simulation with video foundation models for physical {AI}.
\newblock \emph{arXiv preprint arXiv:2511.00062}, 2025.

\bibitem[Du et~al.(2023)Du, Yang, Dai, Dai, Nachum, Tenenbaum, Schuurmans, and
  Abbeel]{unipi}
Y.~Du, S.~Yang, B.~Dai, H.~Dai, O.~Nachum, J.~Tenenbaum, D.~Schuurmans, and
  P.~Abbeel.
\newblock Learning universal policies via text-guided video generation.
\newblock In \emph{Advances in neural information processing systems},
  volume~36, pages 9156--9172, 2023.

\bibitem[Yang et~al.(2024)Yang, Du, Ghasemipour, Tompson, Kaelbling,
  Schuurmans, and Abbeel]{unisim}
S.~Yang, Y.~Du, S.~K.~S. Ghasemipour, J.~Tompson, L.~P. Kaelbling,
  D.~Schuurmans, and P.~Abbeel.
\newblock Learning interactive real-world simulators.
\newblock In \emph{International Conference on Learning Representations}, 2024.

\bibitem[Wu et~al.(2024)Wu, Jing, Cheang, Chen, Xu, Li, Liu, Li, and Kong]{gr1}
H.~Wu, Y.~Jing, C.~Cheang, G.~Chen, J.~Xu, X.~Li, M.~Liu, H.~Li, and T.~Kong.
\newblock Unleashing large-scale video generative pre-training for visual robot
  manipulation.
\newblock In \emph{International Conference on Learning Representations}, 2024.

\bibitem[Hu et~al.(2025)Hu, Guo, Wang, Chen, Wang, Zhang, Sreenath, Lu, and
  Chen]{vpp}
Y.~Hu, Y.~Guo, P.~Wang, X.~Chen, Y.-J. Wang, J.~Zhang, K.~Sreenath, C.~Lu, and
  J.~Chen.
\newblock Video prediction policy: A generalist robot policy with predictive
  visual representations.
\newblock In \emph{International Conference on Machine Learning}, pages
  24328--24346. PMLR, 2025.

\bibitem[Zhu et~al.(2025)Zhu, Wu, Guo, Liu, Cheang, and Kong]{irasim}
F.~Zhu, H.~Wu, S.~Guo, Y.~Liu, C.~Cheang, and T.~Kong.
\newblock Irasim: A fine-grained world model for robot manipulation.
\newblock \emph{Proceedings of the IEEE/CVF International Conference on
  Computer Vision}, pages 9834--9844, 2025.

\bibitem[Ye et~al.(2026)Ye, Ge, Zheng, Gao, Yu, Kurian, Indupuru, Tan, Zhu,
  Xiang, et~al.]{dreamzero}
S.~Ye, Y.~Ge, K.~Zheng, S.~Gao, S.~Yu, G.~Kurian, S.~Indupuru, Y.~L. Tan,
  C.~Zhu, J.~Xiang, et~al.
\newblock World action models are zero-shot policies.
\newblock In \emph{ICLR 2026 the 2nd Workshop on World Models: Understanding,
  Modelling and Scaling}, 2026.

\bibitem[Jang et~al.(2025)Jang, Ye, Lin, Xiang, Bjorck, Fang, Hu, Huang,
  Kundalia, Lin, et~al.]{dreamgen}
J.~Jang, S.~Ye, Z.~Lin, J.~Xiang, J.~Bjorck, Y.~Fang, F.~Hu, S.~Huang,
  K.~Kundalia, Y.-C. Lin, et~al.
\newblock Dreamgen: Unlocking generalization in robot learning through video
  world models.
\newblock \emph{Conference on Robot Learning}, pages 5170--5194, 2025.

\bibitem[Jiang et~al.(2025)Jiang, Chen, Huang, Chen, Zhou, Liao, HE, Liu, Li,
  Yao, et~al.]{enerverseac}
Y.~Jiang, S.~Chen, S.~Huang, L.~Chen, P.~Zhou, Y.~Liao, X.~HE, C.~Liu, H.~Li,
  M.~Yao, et~al.
\newblock Enerverse-ac: Envisioning embodied environments with action
  condition.
\newblock \emph{NeurIPS 2025 Workshop on Embodied World Models for Decision
  Making}, 2025.

\bibitem[Guo et~al.(2026)Guo, Shi, Chen, and Finn]{ctrlworld}
Y.~Guo, L.~X. Shi, J.~Chen, and C.~Finn.
\newblock {Ctrl-World}: A controllable generative world model for robot
  manipulation.
\newblock In \emph{International Conference on Learning Representations
  (ICLR)}, 2026.

\bibitem[Gao et~al.(2025)Gao, Zhou, Du, Zhang, and Gan]{adaworld}
S.~Gao, S.~Zhou, Y.~Du, J.~Zhang, and C.~Gan.
\newblock Adaworld: Learning adaptable world models with latent actions.
\newblock \emph{International Conference on Machine Learning}, pages
  18744--18771, 2025.

\bibitem[Cheang et~al.(2024)Cheang, Chen, Jing, Kong, Li, Li, Liu, Wu, Xu,
  Yang, et~al.]{gr2}
C.-L. Cheang, G.~Chen, Y.~Jing, T.~Kong, H.~Li, Y.~Li, Y.~Liu, H.~Wu, J.~Xu,
  Y.~Yang, et~al.
\newblock Gr-2: A generative video-language-action model with web-scale
  knowledge for robot manipulation.
\newblock \emph{arXiv preprint arXiv:2410.06158}, 2024.

\bibitem[Gao et~al.(2026)Gao, Liang, Zheng, Malik, Ye, Yu, Tseng, Dong, Mo,
  Lin, et~al.]{dreamdojo}
S.~Gao, W.~Liang, K.~Zheng, A.~Malik, S.~Ye, S.~Yu, W.-C. Tseng, Y.~Dong,
  K.~Mo, C.-H. Lin, et~al.
\newblock Dreamdojo: A generalist robot world model from large-scale human
  videos.
\newblock \emph{arXiv preprint arXiv:2602.06949}, 2026.

\bibitem[Bharadhwaj et~al.(2024)Bharadhwaj, Mottaghi, Gupta, and
  Tulsiani]{track2act}
H.~Bharadhwaj, R.~Mottaghi, A.~Gupta, and S.~Tulsiani.
\newblock Track2act: Predicting point tracks from internet videos enables
  generalizable robot manipulation.
\newblock In \emph{European Conference on Computer Vision}, pages 306--324.
  Springer, 2024.

\bibitem[Yang et~al.(2026)Yang, Li, Xu, Wang, Ye, Chen, Qin, Du, Jin, Zhao, and
  Zhao]{orv}
X.~Yang, B.~Li, S.~Xu, N.~Wang, C.~Ye, Z.~Chen, M.~Qin, Y.~Du, X.~Jin, H.~Zhao,
  and H.~Zhao.
\newblock {ORV}: {4D} occupancy-centric robot video generation.
\newblock In \emph{Proceedings of the IEEE/CVF Conference on Computer Vision
  and Pattern Recognition (CVPR)}, 2026.

\bibitem[Wang et~al.(2025)Wang, Wen, Guo, Peng, Qin, Bao, Zhou, and Hu]{vap}
Y.~Wang, C.~Wen, H.~Guo, S.~Peng, M.~Qin, H.~Bao, X.~Zhou, and R.~Hu.
\newblock Precise action-to-video generation through visual action prompts.
\newblock \emph{Proceedings of the IEEE/CVF International Conference on
  Computer Vision}, pages 12713--12724, 2025.

\bibitem[Wan et~al.(2025)Wan, Wang, Ai, Wen, Mao, Xie, Chen, Yu, Zhao, Yang,
  et~al.]{wan21}
T.~Wan, A.~Wang, B.~Ai, B.~Wen, C.~Mao, C.-W. Xie, D.~Chen, F.~Yu, H.~Zhao,
  J.~Yang, et~al.
\newblock Wan: Open and advanced large-scale video generative models.
\newblock \emph{arXiv preprint arXiv:2503.20314}, 2025.

\bibitem[Zhen et~al.(2025)Zhen, Sun, Zhang, Li, Zhou, Du, and Gan]{tessaract}
H.~Zhen, Q.~Sun, H.~Zhang, J.~Li, S.~Zhou, Y.~Du, and C.~Gan.
\newblock {TesserAct}: Learning {4D} embodied world models.
\newblock In \emph{Proceedings of the IEEE/CVF International Conference on
  Computer Vision (ICCV)}, 2025.

\bibitem[Li et~al.(2025{\natexlab{a}})Li, Hsu, Gu, Mees, Pertsch, Walke, Fu,
  Lunawat, Sieh, Kirmani, et~al.]{simpler}
X.~Li, K.~Hsu, J.~Gu, O.~Mees, K.~Pertsch, H.~R. Walke, C.~Fu, I.~Lunawat,
  I.~Sieh, S.~Kirmani, et~al.
\newblock Evaluating real-world robot manipulation policies in simulation.
\newblock In \emph{Conference on Robot Learning}, pages 3705--3728. PMLR,
  2025{\natexlab{a}}.

\bibitem[Li et~al.(2025{\natexlab{b}})Li, Zhu, Wen, Shen, and Xu]{worldeval}
Y.~Li, Y.~Zhu, J.~Wen, C.~Shen, and Y.~Xu.
\newblock Worldeval: World model as real-world robot policies evaluator.
\newblock \emph{arXiv preprint arXiv:2505.19017}, 2025{\natexlab{b}}.

\bibitem[Quevedo et~al.(2025)Quevedo, Sharma, Sun, Suryavanshi, Liang, and
  Yang]{worldgym}
J.~Quevedo, A.~K. Sharma, Y.~Sun, V.~Suryavanshi, P.~Liang, and S.~Yang.
\newblock Worldgym: World model as an environment for policy evaluation.
\newblock \emph{arXiv preprint arXiv:2506.00613}, 2025.

\bibitem[Tseng et~al.(2025)Tseng, Gu, Zhang, Mao, Liu, Shkurti, and
  Yen-Chen]{scalablepolicyeval}
W.-C. Tseng, J.~Gu, Q.~Zhang, H.~Mao, M.-Y. Liu, F.~Shkurti, and L.~Yen-Chen.
\newblock Scalable policy evaluation with video world models.
\newblock \emph{arXiv preprint arXiv:2511.11520}, 2025.

\bibitem[Yue et~al.(2025)Yue, Huang, Liao, Chen, Zhou, Chen, Yao, and
  Ren]{ewmbench}
H.~Yue, S.~Huang, Y.~Liao, S.~Chen, P.~Zhou, L.~Chen, M.~Yao, and G.~Ren.
\newblock Ewmbench: Evaluating scene, motion, and semantic quality in embodied
  world models.
\newblock \emph{arXiv preprint arXiv:2505.09694}, 2025.

\bibitem[Romero et~al.(2017)Romero, Tzionas, and Black]{mano}
J.~Romero, D.~Tzionas, and M.~J. Black.
\newblock Embodied hands: Modeling and capturing hands and bodies together.
\newblock \emph{ACM Transactions on Graphics (Proc. SIGGRAPH Asia)},
  36\penalty0 (6):\penalty0 245:1--245:17, 2017.

\bibitem[Bu et~al.(2025)Bu, Cai, Chen, Cui, Ding, Feng, Gao, He, Hu, Huang,
  et~al.]{agibotworld}
Q.~Bu, J.~Cai, L.~Chen, X.~Cui, Y.~Ding, S.~Feng, S.~Gao, X.~He, X.~Hu,
  X.~Huang, et~al.
\newblock {AgiBot} world colosseo: A large-scale manipulation platform for
  scalable and intelligent embodied systems.
\newblock \emph{arXiv preprint arXiv:2503.06669}, 2025.

\bibitem[Fang et~al.(2023)Fang, Fang, Tang, Liu, Wang, Zhu, and Lu]{rh20t}
H.-S. Fang, H.~Fang, Z.~Tang, J.~Liu, J.~Wang, H.~Zhu, and C.~Lu.
\newblock {RH20T}: A comprehensive robotic dataset for learning diverse skills
  in one-shot.
\newblock \emph{RSS 2023 Workshop on Learning for Task and Motion Planning},
  2023.

\bibitem[Tian et~al.(2025)Tian, Yang, Xie, Cai, Shi, Gao, Liu, Jiang, Qiu,
  Yuan, et~al.]{interndataa1}
Y.~Tian, Y.~Yang, Y.~Xie, Z.~Cai, X.~Shi, N.~Gao, H.~Liu, X.~Jiang, Z.~Qiu,
  F.~Yuan, et~al.
\newblock Interndata-a1: Pioneering high-fidelity synthetic data for
  pre-training generalist policy.
\newblock \emph{arXiv preprint arXiv:2511.16651}, 2025.

\bibitem[Khazatsky et~al.(2024)Khazatsky, Pertsch, Nair, Balakrishna, Dasari,
  Karamcheti, Nasiriany, Srirama, Chen, Ellis, et~al.]{droid}
A.~Khazatsky, K.~Pertsch, S.~Nair, A.~Balakrishna, S.~Dasari, S.~Karamcheti,
  S.~Nasiriany, M.~K. Srirama, L.~Y. Chen, K.~Ellis, et~al.
\newblock {DROID}: A large-scale in-the-wild robot manipulation dataset.
\newblock \emph{arXiv preprint arXiv:2403.12945}, 2024.

\bibitem[Takanami et~al.(2025)Takanami, Khrapchenkov, Morikuni, Arima, Takaba,
  Maeda, Okubo, Sano, Sekioka, Kadoya, et~al.]{airoamoma}
R.~Takanami, P.~Khrapchenkov, S.~Morikuni, J.~Arima, Y.~Takaba, S.~Maeda,
  T.~Okubo, G.~Sano, S.~Sekioka, A.~Kadoya, et~al.
\newblock Airoa moma dataset: A large-scale hierarchical dataset for mobile
  manipulation.
\newblock \emph{arXiv preprint arXiv:2509.25032}, 2025.

\bibitem[Hoque et~al.(2025)Hoque, Huang, Yoon, Sivapurapu, and Zhang]{egodex}
R.~Hoque, P.~Huang, D.~J. Yoon, M.~Sivapurapu, and J.~Zhang.
\newblock Egodex: Learning dexterous manipulation from large-scale egocentric
  video.
\newblock \emph{arXiv preprint arXiv:2505.11709}, 2025.

\bibitem[Li et~al.(2025)Li, Deng, Liang, Luo, Zhou, Yao, Zeng, Feng, Liang, Xu,
  et~al.]{vitra}
Q.~Li, Y.~Deng, Y.~Liang, L.~Luo, L.~Zhou, C.~Yao, L.~Zeng, Z.~Feng, H.~Liang,
  S.~Xu, et~al.
\newblock Scalable vision-language-action model pretraining for robotic
  manipulation with real-life human activity videos.
\newblock \emph{arXiv preprint arXiv:2510.21571}, 2025.

\bibitem[Damen et~al.(2020)Damen, Doughty, Farinella, Fidler, Furnari, Kazakos,
  Moltisanti, Munro, Perrett, Price, et~al.]{damen2020epic}
D.~Damen, H.~Doughty, G.~M. Farinella, S.~Fidler, A.~Furnari, E.~Kazakos,
  D.~Moltisanti, J.~Munro, T.~Perrett, W.~Price, et~al.
\newblock The epic-kitchens dataset: Collection, challenges and baselines.
\newblock \emph{IEEE Transactions on Pattern Analysis and Machine
  Intelligence}, 43\penalty0 (11):\penalty0 4125--4141, 2020.

\bibitem[Zhai et~al.(2023)Zhai, Mustafa, Kolesnikov, and Beyer]{siglip}
X.~Zhai, B.~Mustafa, A.~Kolesnikov, and L.~Beyer.
\newblock Sigmoid loss for language image pre-training.
\newblock In \emph{Proceedings of the IEEE/CVF International Conference on
  Computer Vision (ICCV)}, pages 11975--11986, 2023.

\bibitem[Bai et~al.(2025)Bai, Cai, Chen, Chen, Chen, Cheng, Deng, Ding, Gao,
  Ge, et~al.]{qwen3vl}
S.~Bai, Y.~Cai, R.~Chen, K.~Chen, X.~Chen, Z.~Cheng, L.~Deng, W.~Ding, C.~Gao,
  C.~Ge, et~al.
\newblock Qwen3-vl technical report.
\newblock \emph{arXiv preprint arXiv:2511.21631}, 2025.

\bibitem[Horé and Ziou(2010)]{PSNR}
A.~Horé and D.~Ziou.
\newblock Image quality metrics: Psnr vs. ssim.
\newblock In \emph{2010 20th International Conference on Pattern Recognition},
  pages 2366--2369, 2010.

\bibitem[Wang et~al.(2004)Wang, Bovik, Sheikh, and Simoncelli]{SSIM}
Z.~Wang, A.~Bovik, H.~Sheikh, and E.~Simoncelli.
\newblock Image quality assessment: from error visibility to structural
  similarity.
\newblock \emph{IEEE Transactions on Image Processing}, 13\penalty0
  (4):\penalty0 600--612, 2004.

\bibitem[Zhang et~al.(2018)Zhang, Isola, Efros, Shechtman, and Wang]{lpips}
R.~Zhang, P.~Isola, A.~A. Efros, E.~Shechtman, and O.~Wang.
\newblock The unreasonable effectiveness of deep features as a perceptual
  metric.
\newblock In \emph{Proceedings of the IEEE/CVF Conference on Computer Vision
  and Pattern Recognition (CVPR)}, 2018.

\bibitem[Unterthiner et~al.(2018)Unterthiner, van Steenkiste, Kurach, Marinier,
  Michalski, and Gelly]{fvd}
T.~Unterthiner, S.~van Steenkiste, K.~Kurach, R.~Marinier, M.~Michalski, and
  S.~Gelly.
\newblock Towards accurate generative models of video: A new metric \&
  challenges.
\newblock \emph{arXiv preprint arXiv:1812.01717}, 2018.

\bibitem[Heusel et~al.(2017)Heusel, Ramsauer, Unterthiner, Nessler, and
  Hochreiter]{fid}
M.~Heusel, H.~Ramsauer, T.~Unterthiner, B.~Nessler, and S.~Hochreiter.
\newblock {GANs} trained by a two time-scale update rule converge to a local
  {Nash} equilibrium.
\newblock In \emph{Advances in Neural Information Processing Systems
  (NeurIPS)}, 2017.

\bibitem[Wang et~al.(2026)Wang, Xu, Dong, Deng, Xiang, Lv, Sun, Tong, and
  Yang]{mogev2}
R.~Wang, S.~Xu, Y.~Dong, Y.~Deng, J.~Xiang, Z.~Lv, G.~Sun, X.~Tong, and
  J.~Yang.
\newblock Moge-2: Accurate monocular geometry with metric scale and sharp
  details.
\newblock \emph{Advances in Neural Information Processing Systems},
  38:\penalty0 35928--35959, 2026.

\bibitem[Lu et~al.(2025)Lu, Liang, Xie, Richter, Lin, Liu, and Yip]{ctrnetx}
J.~Lu, Z.~Liang, T.~Xie, F.~Richter, S.~Lin, S.~Liu, and M.~C. Yip.
\newblock Ctrnet-x: Camera-to-robot pose estimation in real-world conditions
  using a single camera.
\newblock In \emph{2025 IEEE International Conference on Robotics and
  Automation (ICRA)}, pages 1914--1920. IEEE, 2025.

\bibitem[Kwon et~al.(2023)Kwon, Li, Zhuang, Sheng, Zheng, Yu, Gonzalez, Zhang,
  and Stoica]{vllm}
W.~Kwon, Z.~Li, S.~Zhuang, Y.~Sheng, L.~Zheng, C.~H. Yu, J.~E. Gonzalez,
  H.~Zhang, and I.~Stoica.
\newblock Efficient memory management for large language model serving with
  {PagedAttention}.
\newblock In \emph{Proceedings of the 29th Symposium on Operating Systems
  Principles (SOSP)}, pages 611--626, 2023.

\bibitem[Loshchilov and Hutter(2018)]{adamw}
I.~Loshchilov and F.~Hutter.
\newblock Decoupled weight decay regularization.
\newblock In \emph{International Conference on Learning Representations}, 2018.

\bibitem[Ho and Salimans(2022)]{ho2022classifier}
J.~Ho and T.~Salimans.
\newblock Classifier-free diffusion guidance.
\newblock \emph{NeurIPS 2021 Workshop on Deep Generative Models and Downstream
  Applications}, 2022.

\bibitem[Grauman et~al.(2022)Grauman, Westbury, Byrne, Chavis, Furnari,
  Girdhar, Hamburger, Jiang, Liu, Liu, et~al.]{ego4d}
K.~Grauman, A.~Westbury, E.~Byrne, Z.~Chavis, A.~Furnari, R.~Girdhar,
  J.~Hamburger, H.~Jiang, M.~Liu, X.~Liu, et~al.
\newblock Ego4d: Around the world in 3,000 hours of egocentric video.
\newblock In \emph{Proceedings of the IEEE/CVF Conference on Computer Vision
  and Pattern Recognition (CVPR)}, pages 18995--19012, 2022.

\end{thebibliography}
}

\clearpage


\appendix

\section{Technical Appendix}
\subsection{Data curation pipeline}
\subsubsection{Dataset details}
\label{app:dataset_details}

The main paper (\S\ref{sec:data_sources}) summarizes dataset scale in Table~\ref{tab:data_sources}. Here we provide the per-source collection details and dataset-specific notes.

\paragraph{Robot sources.}
The robot sources cover distinct scene types and tasks. RH20T (cfg5 and cfg7) records contact-rich tabletop manipulation on Franka and KUKA, covering 147 tasks across 42 skills such as cutting, pouring, folding, and assembly. AIROA-MoMa records mobile manipulation on a Toyota HSR, where the wheeled base lets episodes leave the tabletop and operate across rooms. DROID covers 86 tasks across 564 real-world scenes. AgiBot World stages 217 tasks across 87 skills on the AgiBot G1 humanoid in five settings: domestic, retail, industrial, restaurant, and office. InternData-A1 is synthetic and renders 70 tasks across 18 skills in 227 indoor scenes.

\paragraph{Human sources.}
The two human sources differ from the robot sources in collection mechanism: a head-mounted camera follows the operator across rooms and homes, with no fixed rig to instrument or relocate. EgoDex uses Apple Vision Pro to record 194 everyday tabletop tasks, ranging from tying shoelaces to folding laundry. EPIC-Kitchens captures unscripted daily cooking in 45 real home kitchens across 4 cities. Because head-mounted cameras are much easier to set up than robot rigs, a single operator can record across many real homes and kitchens. 

\paragraph{Notes for Table~\ref{tab:data_sources}.}
\phantomsection\label{note:synthetic}$^{1}$ InternData-A1 is synthetic; all other sources are real-world recordings.
\phantomsection\label{note:rh20t}$^{2}$ RH20T spans seven configurations. We restrict to cfg5 (Franka) and cfg7 (KUKA) because cfg1 lacks proprioceptive state and cfg3 has camera-extrinsic miscalibration producing large skeleton--image offsets. Released hours are estimated from cfg5+cfg7 episode counts since \citet{rh20t} does not publish per-configuration hours. Mean episode duration $\approx 40$\,s comes from the Fig.~3 histogram.

\subsubsection{Captioning prompt}
\label{app:caption_prompt}

We caption every retained episode with Qwen3-VL-30B-A3B-Instruct~\citep{qwen3vl} served by vLLM~\citep{vllm}; see~\S\ref{sec:data_captioning} for the captioning procedure. Robot and human episodes use the identical system prompt shown below.

\Needspace{14\baselineskip}
\begin{tcolorbox}[
  breakable,
  colback=black!2,
  colframe=black!35,
  boxrule=0.4pt,
  arc=1mm,
  left=1mm,right=1mm,top=0.8mm,bottom=0.8mm,
  title={System prompt used for captioning},
  fonttitle=\bfseries\footnotesize,
  fontupper=\ttfamily\footnotesize,
]
\raggedright\sloppy
You are a video captioning specialist whose goal is to generate high-quality English prompts by referring to the details of the user's input videos. Your task is to carefully analyze the content, context, and actions within the video, and produce a complete, factual, precise and natural-sounding caption that accurately describes the video. The caption should preserve the original intent and meaning of the video while maintaining the clarity and original meaning, strictly adhering to the formatting of the examples provided.

\medskip
\textbf{Task Requirements:}
\begin{enumerate}
  \setlength{\itemsep}{0pt}
  \setlength{\parskip}{0pt}
  \setlength{\parsep}{0pt}
  \item You need to describe the main subject of the video in detail, including their appearance, actions, expressions, and surrounding environment.
  \item You need to emphasize movement information in the input and different camera angles.
  \item Your output should convey natural movement attributes, incorporating natural actions related to the described subject category, using simple and direct verbs as much as possible.
  \item You should reference the detailed information in the video, such as character actions, clothing, backgrounds, and emphasize the details in the video.
  \item Control the rewritten prompt to around 80--100 words.
  \item No matter what language the user inputs, you must always output in English.
\end{enumerate}

\medskip
\textbf{Example of the rewritten English prompt:}
\begin{enumerate}
  \setlength{\itemsep}{2pt}
  \setlength{\parskip}{0pt}
  \setlength{\parsep}{0pt}
  \item A Japanese fresh film-style photo of a young East Asian girl with double braids sitting by the boat. The girl wears a white square collar puff sleeve dress, decorated with pleats and buttons. She has fair skin, delicate features, and slightly melancholic eyes, staring directly at the camera. Her hair falls naturally, with bangs covering part of her forehead. She rests her hands on the boat, appearing natural and relaxed. The background features a blurred outdoor scene, with hints of blue sky, mountains, and some dry plants. The photo has a vintage film texture. A medium shot of a seated portrait.
  \item An anime illustration in vibrant thick painting style of a white girl with cat ears holding a folder, showing a slightly dissatisfied expression. She has long dark purple hair and red eyes, wearing a dark gray skirt and a light gray top with a white waist tie and a name tag in bold Chinese characters that says ``Ziyang''. The background has a light yellow indoor tone, with faint outlines of some furniture visible. A pink halo hovers above her head, in a smooth Japanese cel-shading style. A close-up shot from a slightly elevated perspective.
  \item CG game concept digital art featuring a huge crocodile with its mouth wide open, with trees and thorns growing on its back. The crocodile's skin is rough and grayish-white, resembling stone or wood texture. Its back is lush with trees, shrubs, and thorny protrusions. With its mouth agape, the crocodile reveals a pink tongue and sharp teeth. The background features a dusk sky with some distant trees, giving the overall scene a dark and cold atmosphere. A close-up from a low angle.
  \item In the style of an American drama promotional poster, Walter White sits in a metal folding chair wearing a yellow protective suit, with the words ``Breaking Bad'' written in sans-serif English above him, surrounded by piles of dollar bills and blue plastic storage boxes. He wears glasses, staring forward, dressed in a yellow jumpsuit, with his hands resting on his knees, exuding a calm and confident demeanor. The background shows an abandoned, dim factory with light filtering through the windows. There's a noticeable grainy texture. A medium shot with a straight-on close-up of the character.
\end{enumerate}

\medskip
Directly output the rewritten English text.
\end{tcolorbox}

\subsection{Evaluation benchmark selection}
\label{app:eval_benchmark}

The $200$-clip evaluation benchmark of \S\ref{sec:exp_setup} is curated from six robot datasets by two rules. First, we run k-means in the joint space of caption embeddings and end-effector motion magnitude to remove near-duplicate scenes. Second, we enforce arm visibility with a forward-window check: a clip is kept only if the end effector remains within the camera frame for the full evaluation horizon. We also impose per-embodiment quotas proportional to $\sqrt{N}$ of each source corpus, balancing the long-tail KUKA and Franka data against AgiBot. 

\subsection{Training implementation details}
\label{app:training_setup}

We finetune from the pretrained Cosmos-Predict2.5-2B~\citep{cosmos-predict2p5} checkpoint with AdamW~\citep{adamw} at learning rate $3{\times}10^{-5}$ and batch size $16$. Timesteps are sampled from a logit-normal distribution with reweighting, and we use shift parameter $5$. To balance different embodiment sources, we draw batches with frequency-tempered weights $w_i \propto n_{\mathrm{frames},i}^{1/T}$ with $T{=}3$, which upweights smaller sources without tuning per-source coefficients. For each $81$-frame training window, we choose the start frame with a bias toward grasp and release events: the start index is drawn from a trapezoid prior over midpoint crossings of a binary open/close signal (gripper openness for robots; normalised fingertip flexion for humans); During training, classifier-free guidance~\citep{ho2022classifier} replaces $S_{1:T}$ with zeros with probability $0.2$; at inference we use guidance scale $w{=}6$. The two-stage schedule trains $15$k iterations on the four robot embodiments and then continues on the full robot plus human data mixture.

\subsection{Latent-action conditioning pathway}
\label{app:latent_action_impl}

The ``Latent Action'' row in Table~\ref{tab:abl} uses a latent-action conditioning baseline. For reproducibility, we detail the conditioning pipeline here. Our implementation follows the Cosmos-Predict2.5~\cite{cosmos-predict2p5} latent action conditioned generation; we report only the components required to integrate it with our diffusion backbone.

\paragraph{Latent action.} Each arm is represented by a $7$-D action consisting of: $3$ end-effector translation, $3$ rotation (Euler angles), and $1$ gripper-openness scalar. We convert states to actions by taking frame-to-frame differences expressed in the previous frame's local coordinate system. Concretely, the translation delta is rotated by the previous orientation. The rotation delta is computed via the relative rotation $R_{t-1}^\top R_t$, which is then converted to Euler angles. We scale both translation and rotation deltas by $20$, and keep the gripper value unchanged.

\paragraph{Multi-embodiment alignment.} We adopt a single, fixed $14$-D action vector across all robot embodiments. For bimanual robots (AgiBot G1), we concatenate the left- and right-arm $7$-D states. For single-arm robots (Franka Panda, KUKA iiwa, Toyota HSR), we place the arm state in the first $7$ dimensions and set the remaining $7$ dimensions to zero. This yields a unified interface and avoids embodiment-specific prediction heads.

\paragraph{Token projection.} Each clip contains $T=81$ frames and therefore $T-1=80$ transitions. The resulting action tensor has shape $(80,14)$. We flatten the $(80,14)$ action tensor into a $1120$-D vector and project it into conditioning tokens using two MLPs. Each MLP has one GELU-activated hidden layer of width $4D$ followed by a linear output layer. One MLP produces a $D$-dimensional token, and the other produces a $3D$-dimensional token. We set $D=2048$, matching the DiT hidden size.

\paragraph{Injection points.} The $D$-dimensional token is added to the timestep embedding at every frame. The $3D$-dimensional token is added to the adaptive LayerNorm modulation signal at every frame; in each DiT block this signal is split into shift, scale, and gate vectors that modulate the block's LayerNorm. This gives a single clip-level action signal shared across all frames, and adds no extra spatial tokens.

\subsection{Baseline configurations}
\label{app:baselines}
Table~\ref{tab:baselines} provides a detailed comparison of our baselines, including model size, conditioning signals, and training data sources.

\begin{table}[t]
\centering
\caption{\footnotesize Baseline configurations: parameter count, conditioning representations, and training data.
}
\label{tab:baselines}
\small
\setlength{\tabcolsep}{3pt}
\renewcommand{\arraystretch}{1.05}
\begin{tabular}{lcl>{\raggedright\arraybackslash}p{6.7cm}}
\toprule
Method & Size & Condition & Training Data \\
\midrule
Cosmos-Predict2.5~\citep{cosmos-predict2p5} & 2B  & Text           & Bridge, AgiBot World, LIBERO, RoboCasa \\
TesserAct~\citep{tessaract}                      & 5B  & Text           & RT-1, Bridge, RLBench, SS-v2 \\
\addlinespace[2pt]
IRASim~\citep{irasim}                            & 0.7B & Latent Action & RT-1, Bridge, Lang.-Table, RoboNet \\
Ctrl-World~\citep{ctrlworld}                     & 1.5B & Latent Action & DROID \\
\addlinespace[2pt]
EnerVerse-AC~\citep{enerverseac}                 & 1.5B$^{2}$ & Gripper render.\,+\,LA & AgiBot World \\
Genie Envisioner~\citep{genie-envisioner}        & 2B  & Gripper render.          & AgiBot World \\
Kinema4D~\citep{kinema4d}                        & 14B & Point map                & Robo4D-200k$^{3}$ \\
\midrule
\methodname{} (\textbf{Ours})                    & 2B  & Skeleton                 & See Table~\ref{tab:data_sources}. \\
\bottomrule
\end{tabular}
\end{table}

\footnotetext[1]{IRASim is only post-trained on a specific in-distribution dataset, and has little generalization capability.}
\footnotetext[2]{EnerVerse-AC reports no parameter count in any official channel; we report the U-Net trainable count (1.46B) measured from the released DeepSpeed checkpoint.}

\subsection{Per-embodiment quantitative results}
\label{app:per_emb}

Table~\ref{tab:main_quant} reports scores averaged across the six robot subsets in the benchmark. Here, we disaggregate these averages and provide per-embodiment results. Specifically, each table corresponds to a single subset and reports five quality metrics computed by all baselines under the same evaluation protocol as Table~\ref{tab:main_quant} (PSNR, SSIM, LPIPS, tLPIPS, and $\text{L2}_{\text{latent}}$).
\textbf{Best} per column is in \textbf{bold} and \underline{second-best} is underlined. IRASim's $7$-DoF action interface is incompatible with the bimanual AgiBot G1 and the mobile-base AIROA-MoMa subsets, so it is excluded from those two tables.

\begin{table*}[t]
\centering
\setlength{\tabcolsep}{4pt}

\begin{minipage}[t]{0.49\textwidth}
\centering
\caption{\footnotesize Per-embodiment results on \textbf{AgiBot G1} ($N = 75$).}
\label{tab:per_emb_agibot}
\resizebox{\linewidth}{!}{%
\begin{tabular}{lccccc}
\toprule
Method & PSNR$\uparrow$ & SSIM$\uparrow$ & LPIPS$\downarrow$ & tLPIPS$\downarrow$ & $\text{L2}_{\text{latent}}$$\downarrow$ \\
\midrule
Cosmos-Predict2.5~\citep{cosmos-predict2p5}    & 13.12 & 0.468 & 0.491 & 0.020 & 0.576 \\
TesserAct~\citep{tessaract}                    & 15.35 & 0.708 & 0.304 & 0.062 & 0.409 \\
\addlinespace[2pt]
Ctrl-World~\citep{ctrlworld}                   & 19.10 & 0.710 & 0.329 & 0.039 & 0.272 \\
\addlinespace[2pt]
EnerVerse-AC~\citep{enerverseac}               & 21.20 & 0.767 & 0.214 & 0.019 & 0.158 \\
Genie Envisioner~\citep{genie-envisioner}      & \underline{23.33} & \textbf{0.855} & \underline{0.131} & \textbf{0.007} & \underline{0.105} \\
Kinema4D~\citep{kinema4d}                      & 17.29 & 0.742 & 0.193 & 0.018 & 0.223 \\
\midrule
\methodname{} (\textbf{Ours})                  & \textbf{23.70} & \underline{0.852} & \textbf{0.079} & \underline{0.013} & \textbf{0.076} \\
\bottomrule
\end{tabular}}
\end{minipage}\hfill
\begin{minipage}[t]{0.49\textwidth}
\centering
\caption{\footnotesize Per-embodiment results on \textbf{AIROA-MoMa} ($N = 23$).}
\label{tab:per_emb_airoa}
\resizebox{\linewidth}{!}{%
\begin{tabular}{lccccc}
\toprule
Method & PSNR$\uparrow$ & SSIM$\uparrow$ & LPIPS$\downarrow$ & tLPIPS$\downarrow$ & $\text{L2}_{\text{latent}}$$\downarrow$ \\
\midrule
Cosmos-Predict2.5~\citep{cosmos-predict2p5}    & 12.29 & 0.520 & 0.523 & 0.044 & 0.723 \\
TesserAct~\citep{tessaract}                    & 15.56 & 0.686 & 0.341 & 0.066 & 0.483 \\
\addlinespace[2pt]
Ctrl-World~\citep{ctrlworld}                   & 18.60 & 0.699 & 0.349 & 0.056 & 0.397 \\
\addlinespace[2pt]
EnerVerse-AC~\citep{enerverseac}               & 18.41 & 0.675 & 0.317 & \underline{0.018} & 0.369 \\
Genie Envisioner~\citep{genie-envisioner}      & \textbf{23.31} & \textbf{0.785} & \textbf{0.195} & \textbf{0.015} & \textbf{0.208} \\
Kinema4D~\citep{kinema4d}                      & 16.78 & 0.714 & 0.266 & 0.021 & 0.349 \\
\midrule
\methodname{} (\textbf{Ours})                  & \underline{20.77} & \underline{0.715} & \underline{0.254} & 0.035 & \underline{0.312} \\
\bottomrule
\end{tabular}}
\end{minipage}
\end{table*}

\begin{table*}[t]
\centering
\setlength{\tabcolsep}{4pt}

\begin{minipage}[t]{0.49\textwidth}
\centering
\caption{\footnotesize Per-embodiment results on \textbf{DROID} ($N = 29$).}
\label{tab:per_emb_droid}
\resizebox{\linewidth}{!}{%
\begin{tabular}{lccccc}
\toprule
Method & PSNR$\uparrow$ & SSIM$\uparrow$ & LPIPS$\downarrow$ & tLPIPS$\downarrow$ & $\text{L2}_{\text{latent}}$$\downarrow$ \\
\midrule
Cosmos-Predict2.5~\citep{cosmos-predict2p5}    & 16.40 & 0.717 & 0.243 & 0.021 & 0.263 \\
TesserAct~\citep{tessaract}                    & 15.81 & 0.742 & 0.265 & 0.048 & 0.356 \\
\addlinespace[2pt]
IRASim~\citep{irasim}                          & 6.45  & 0.090 & 0.905 & 0.606 & 2.321 \\
Ctrl-World~\citep{ctrlworld}                   & 17.40 & 0.712 & 0.330 & 0.059 & 0.304 \\
\addlinespace[2pt]
EnerVerse-AC~\citep{enerverseac}               & 18.40 & 0.729 & 0.229 & 0.027 & 0.214 \\
Genie Envisioner~\citep{genie-envisioner}      & \underline{20.52} & \underline{0.821} & \underline{0.154} & \textbf{0.005} & \underline{0.182} \\
Kinema4D~\citep{kinema4d}                      & 16.97 & 0.746 & 0.226 & 0.042 & 0.299 \\
\midrule
\methodname{} (\textbf{Ours})                  & \textbf{21.60} & \textbf{0.834} & \textbf{0.102} & \underline{0.020} & \textbf{0.113} \\
\bottomrule
\end{tabular}}%
\end{minipage}\hfill
\begin{minipage}[t]{0.49\textwidth}
\centering
\caption{\footnotesize Per-embodiment results on \textbf{InternData} ($N = 30$).}
\label{tab:per_emb_interndata}
\resizebox{\linewidth}{!}{%
\begin{tabular}{lccccc}
\toprule
Method & PSNR$\uparrow$ & SSIM$\uparrow$ & LPIPS$\downarrow$ & tLPIPS$\downarrow$ & $\text{L2}_{\text{latent}}$$\downarrow$ \\
\midrule
Cosmos-Predict2.5~\citep{cosmos-predict2p5}    & 16.13 & 0.602 & 0.344 & 0.025 & 0.342 \\
TesserAct~\citep{tessaract}                    & 19.95 & 0.814 & 0.210 & 0.045 & 0.209 \\
\addlinespace[2pt]
IRASim~\citep{irasim}                          & 6.48  & 0.096 & 0.997 & 0.611 & 2.504 \\
Ctrl-World~\citep{ctrlworld}                   & 21.42 & 0.732 & 0.305 & 0.031 &  0.282 \\
\addlinespace[2pt]
EnerVerse-AC~\citep{enerverseac}               & 21.54 & 0.758 & 0.233 & 0.017 & 0.192 \\
Genie Envisioner~\citep{genie-envisioner}      & \underline{25.35} & \underline{0.839} & \underline{0.136} & \textbf{0.006} & \underline{0.102} \\
Kinema4D~\citep{kinema4d}                      & 20.85 & 0.787 & 0.138 & 0.011 & 0.121 \\
\midrule
\methodname{} (\textbf{Ours})                  & \textbf{27.65} & \textbf{0.864} & \textbf{0.061} & \underline{0.008} & \textbf{0.042} \\
\bottomrule
\end{tabular}}%
\end{minipage}
\end{table*}

\begin{table*}[t]
\centering
\setlength{\tabcolsep}{4pt}

\begin{minipage}[t]{0.49\textwidth}
\centering
\caption{\footnotesize Per-embodiment results on \textbf{RH20T-cfg5} ($N = 23$).}
\label{tab:per_emb_rh20t_cfg5}
\resizebox{\linewidth}{!}{%
\begin{tabular}{lccccc}
\toprule
Method & PSNR$\uparrow$ & SSIM$\uparrow$ & LPIPS$\downarrow$ & tLPIPS$\downarrow$ & $\text{L2}_{\text{latent}}$$\downarrow$ \\
\midrule
Cosmos-Predict2.5~\citep{cosmos-predict2p5}    & 17.85 & 0.661 & 0.167 & \underline{0.013} & 0.205 \\
TesserAct~\citep{tessaract}                    & 15.77 & 0.715 & 0.285 & 0.053 & 0.385 \\
\addlinespace[2pt]
IRASim~\citep{irasim}                          & 6.76  & 0.088 & 0.859 & 0.602 & 2.603 \\
Ctrl-World~\citep{ctrlworld}                   & 19.08 & 0.705 & 0.292 & 0.035 & 0.270 \\
\addlinespace[2pt]
EnerVerse-AC~\citep{enerverseac}               & 22.12 & 0.775 & 0.156 & 0.020 & 0.152 \\
Genie Envisioner~\citep{genie-envisioner}      & \underline{24.35} & \underline{0.856} & \underline{0.117} & \textbf{0.008} & \underline{0.118} \\
Kinema4D~\citep{kinema4d}                      & 16.28 & 0.707 & 0.219 & 0.025 & 0.268 \\
\midrule
\methodname{} (\textbf{Ours})                  & \textbf{26.92} & \textbf{0.906} & \textbf{0.050} & \textbf{0.008} & \textbf{0.046} \\
\bottomrule
\end{tabular}}%
\end{minipage}\hfill
\begin{minipage}[t]{0.49\textwidth}
\centering
\caption{\footnotesize Per-embodiment results on \textbf{RH20T-cfg7} ($N = 21$).}
\label{tab:per_emb_rh20t_cfg7}
\resizebox{\linewidth}{!}{%
\begin{tabular}{lccccc}
\toprule
Method & PSNR$\uparrow$ & SSIM$\uparrow$ & LPIPS$\downarrow$ & tLPIPS$\downarrow$ & $\text{L2}_{\text{latent}}$$\downarrow$ \\
\midrule
Cosmos-Predict2.5~\citep{cosmos-predict2p5}    & 15.93 & 0.577 & 0.202 & 0.013 & 0.244 \\
TesserAct~\citep{tessaract}                    & 16.18 & 0.741 & 0.218 & 0.039 & 0.284 \\
\addlinespace[2pt]
IRASim~\citep{irasim}                          & 6.23  & 0.072 & 0.841 & 0.601 & 2.396 \\
Ctrl-World~\citep{ctrlworld}                   & 18.30 & 0.649 & 0.302 & 0.041 & 0.268 \\
\addlinespace[2pt]
EnerVerse-AC~\citep{enerverseac}               & 19.64 & 0.719 & 0.201 & 0.032 & 0.184 \\
Genie Envisioner~\citep{genie-envisioner}      & \underline{22.86} & \underline{0.835} & \underline{0.118} & \textbf{0.005} & \underline{0.111} \\
Kinema4D~\citep{kinema4d}                      & 18.08 & 0.729 & 0.165 & 0.016 & 0.175 \\
\midrule
\methodname{} (\textbf{Ours})                  & \textbf{25.85} & \textbf{0.892} & \textbf{0.051} & \underline{0.008} & \textbf{0.041} \\
\bottomrule
\end{tabular}}%
\end{minipage}
\end{table*}

\subsection{Per-embodiment qualitative results}
\label{app:per_emb_qual}

\begin{figure}[t]
\centering
\setlength{\tabcolsep}{2pt}
\renewcommand{\arraystretch}{1.0}
\begin{tabular}{@{}cc@{}}
\includegraphics[width=0.49\linewidth]{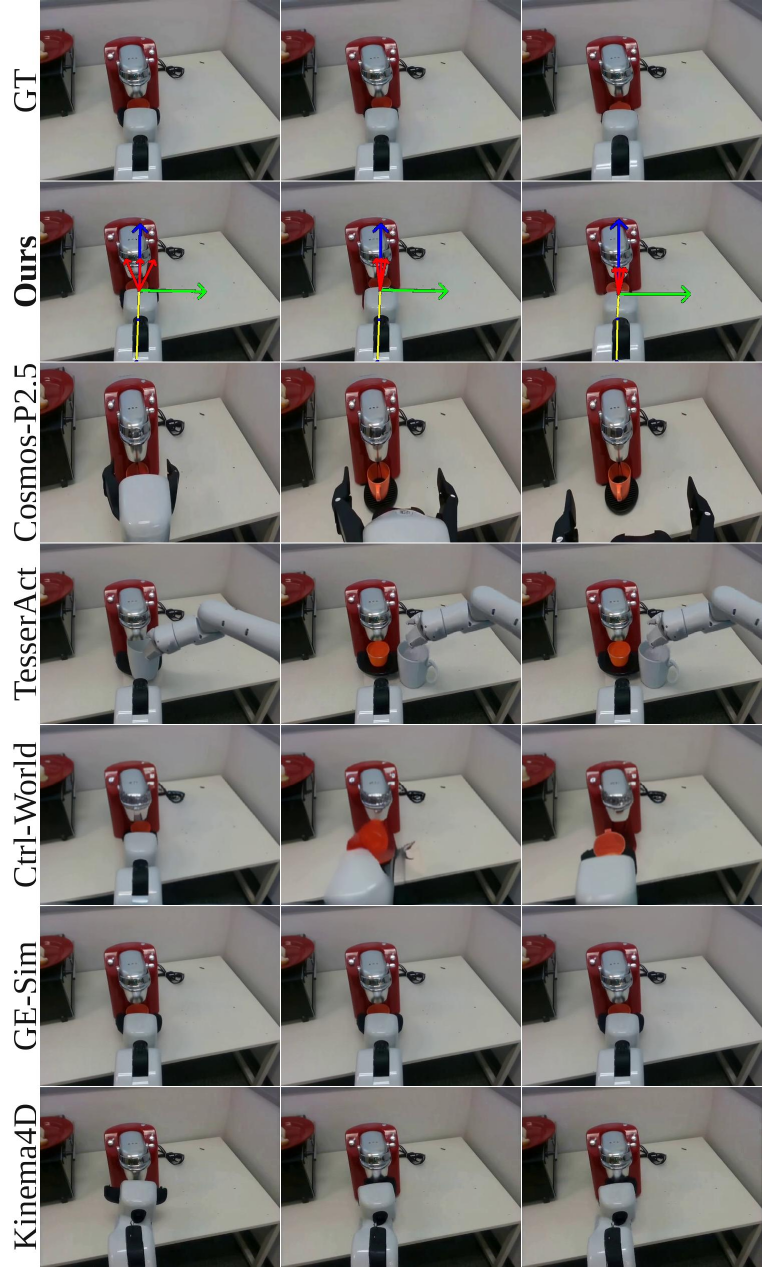} &
\includegraphics[width=0.49\linewidth]{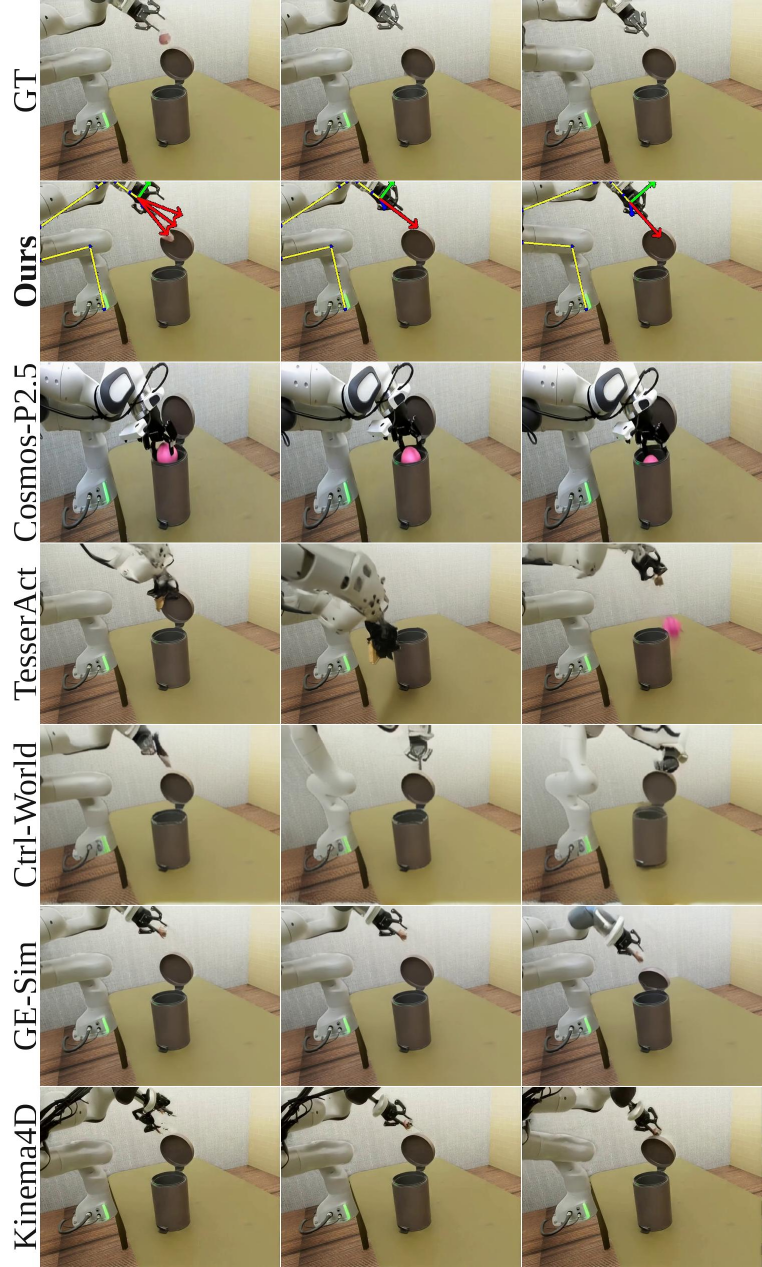} \\
\textbf{\textcolor{cAIROA}{AIROA-MoMa}} &
\textbf{\textcolor{cIntern}{InternData}} \\[2pt]
\includegraphics[width=0.49\linewidth]{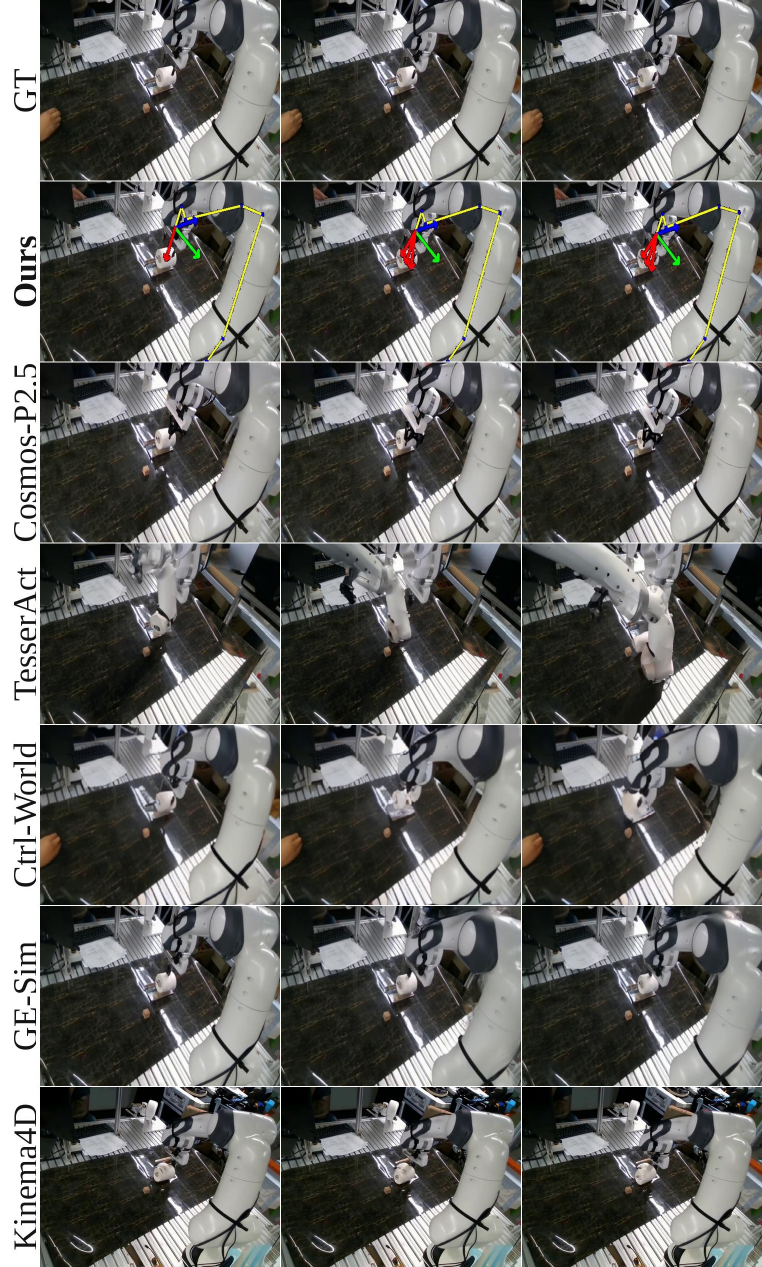} &
\includegraphics[width=0.49\linewidth]{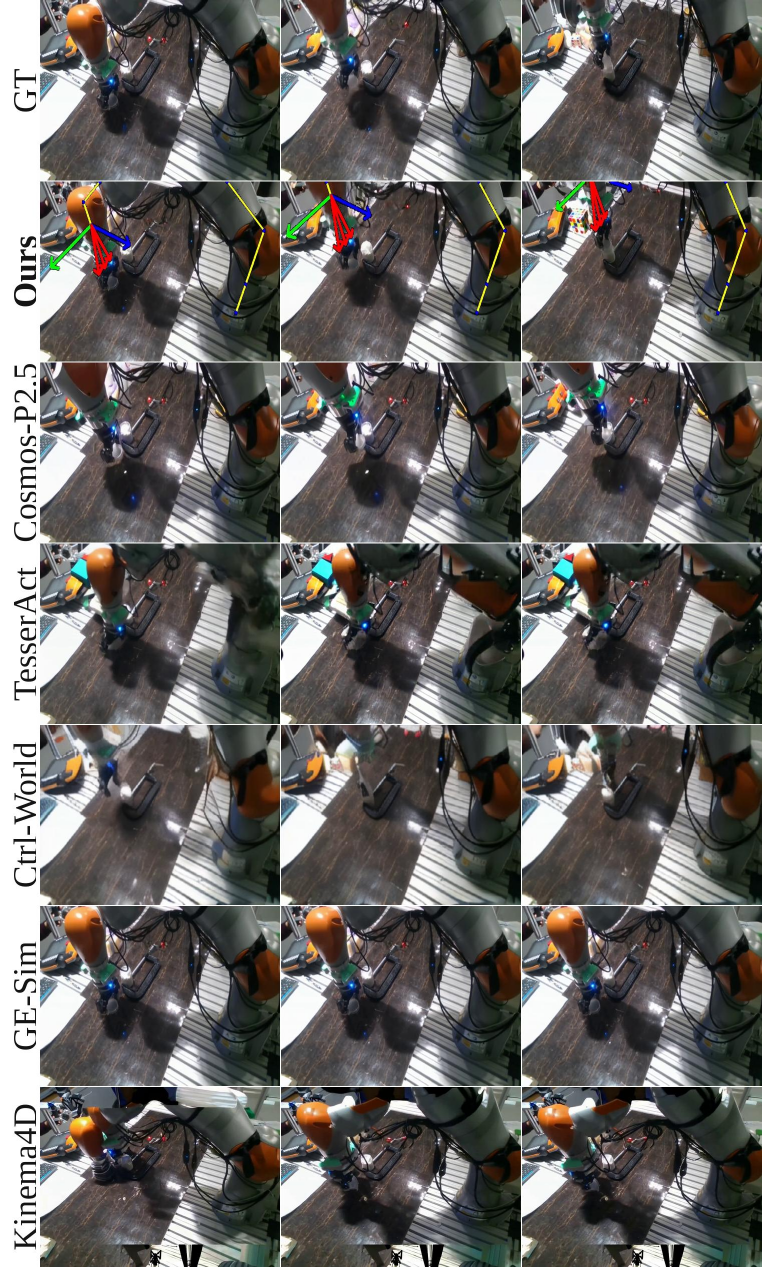} \\
\textbf{\textcolor{cRH20T5}{RH20T-cfg5}} &
\textbf{\textcolor{cRH20T7}{RH20T-cfg7}} \\
\end{tabular}
\caption{Qualitative comparison on the four remaining robot embodiments. Embodiment colours follow Figure~\ref{fig:skeleton_grid}.}
\label{fig:qual_appendix_robots}
\end{figure}

\begin{figure}[t]
\centering
\setlength{\tabcolsep}{2pt}
\renewcommand{\arraystretch}{1.0}
\begin{tabular}{@{}cc@{}}
\includegraphics[width=0.49\linewidth]{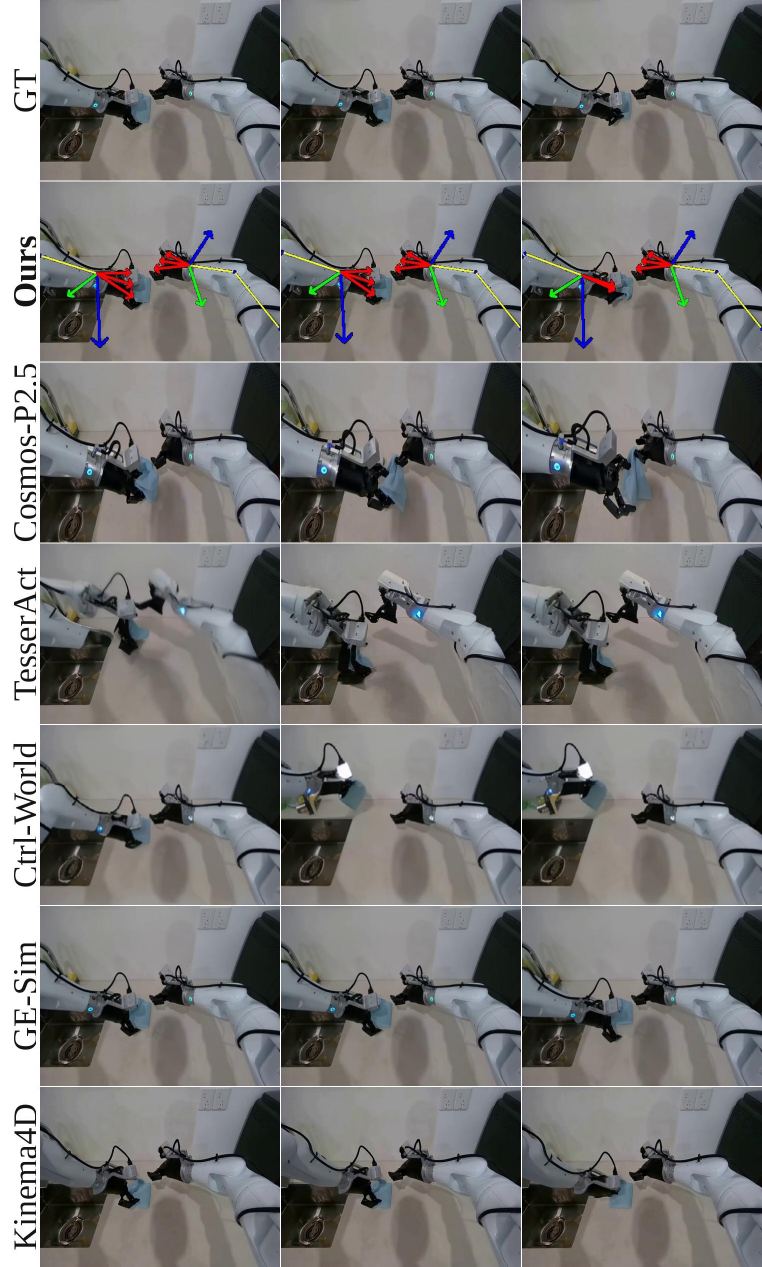} &
\includegraphics[width=0.49\linewidth]{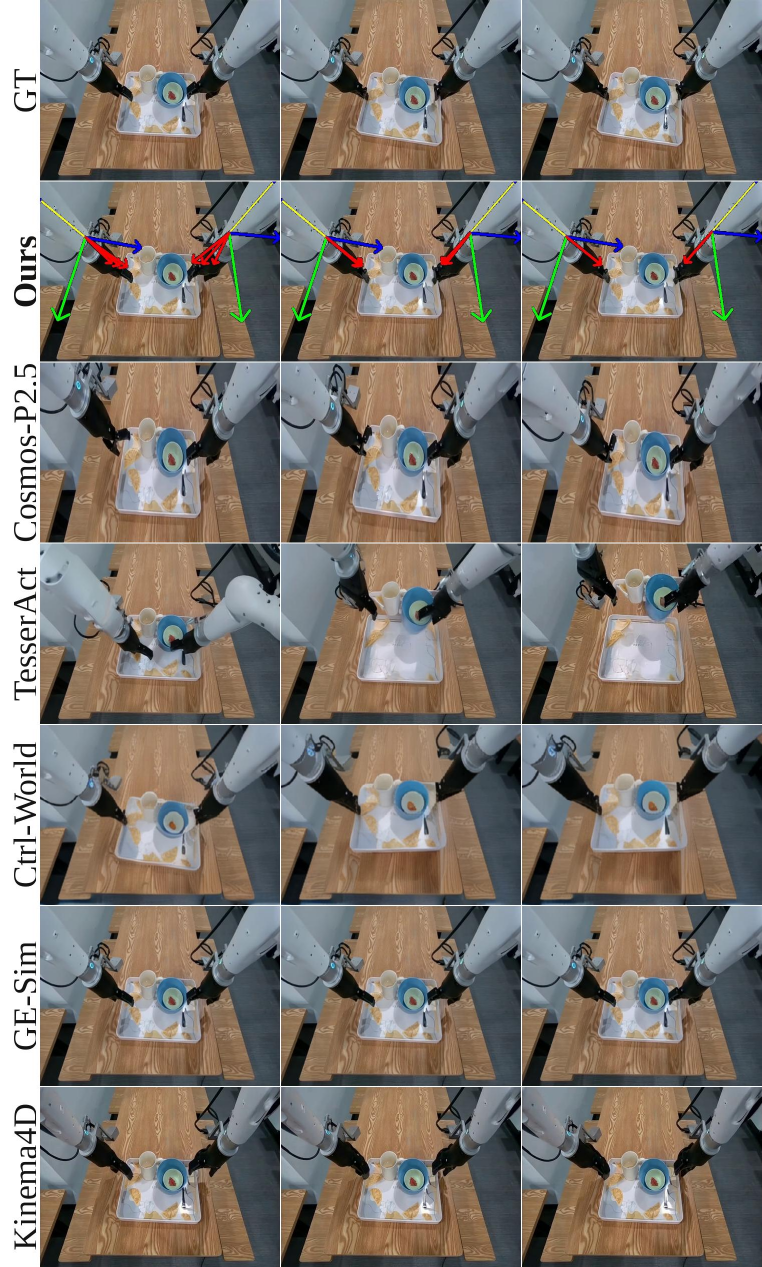} \\
\textbf{\textcolor{cAgiBot}{AgiBot G1}} &
\textbf{\textcolor{cAgiBot}{AgiBot G1}} \\[2pt]
\includegraphics[width=0.49\linewidth]{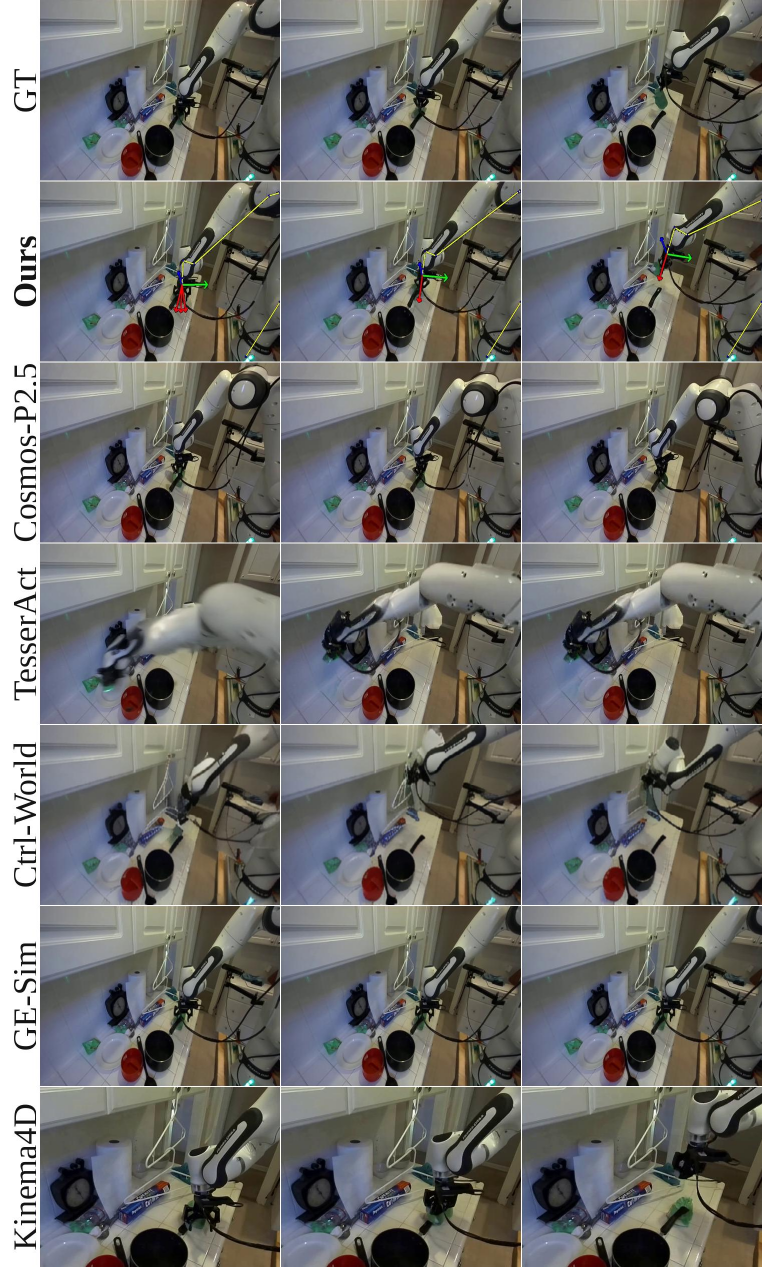} &
\includegraphics[width=0.49\linewidth]{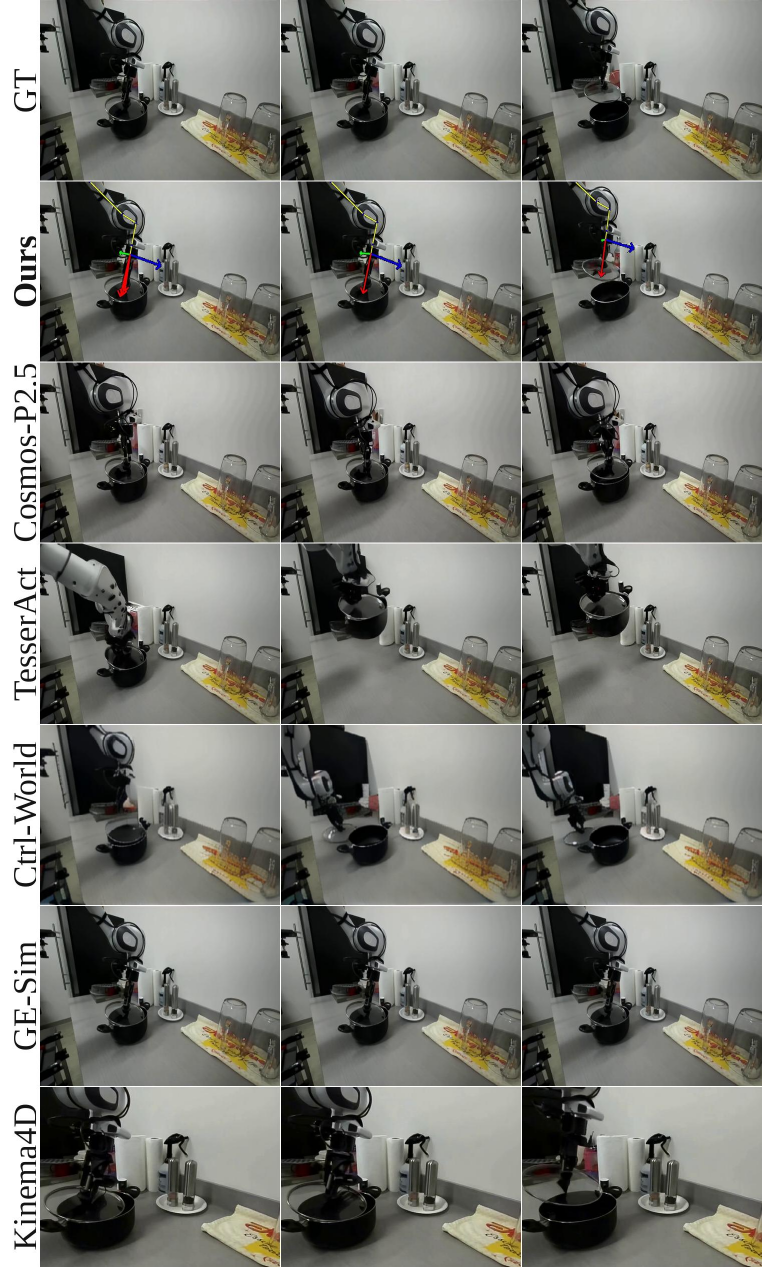} \\
\textbf{\textcolor{cDROID}{DROID}} &
\textbf{\textcolor{cDROID}{DROID}} \\
\end{tabular}
\caption{Additional qualitative samples for AgiBot G1 and DROID, complementing Figure~\ref{fig:qual_compare}.}
\label{fig:qual_appendix_extras}
\end{figure}

Figure~\ref{fig:qual_compare} in the main paper presents qualitative results on two embodiments. Here, we extend the comparison by including the remaining four robot embodiments (Figure~\ref{fig:qual_appendix_robots}), along with two additional examples each for AgiBot G1 and DROID (Figure~\ref{fig:qual_appendix_extras}).

\subsection{Ablation qualitative comparisons}
\label{app:abl_qual}

\begin{figure}[t]
\centering
\setlength{\tabcolsep}{2pt}
\renewcommand{\arraystretch}{1.0}
\begin{tabular}{@{}cc@{}}
\includegraphics[width=0.49\linewidth]{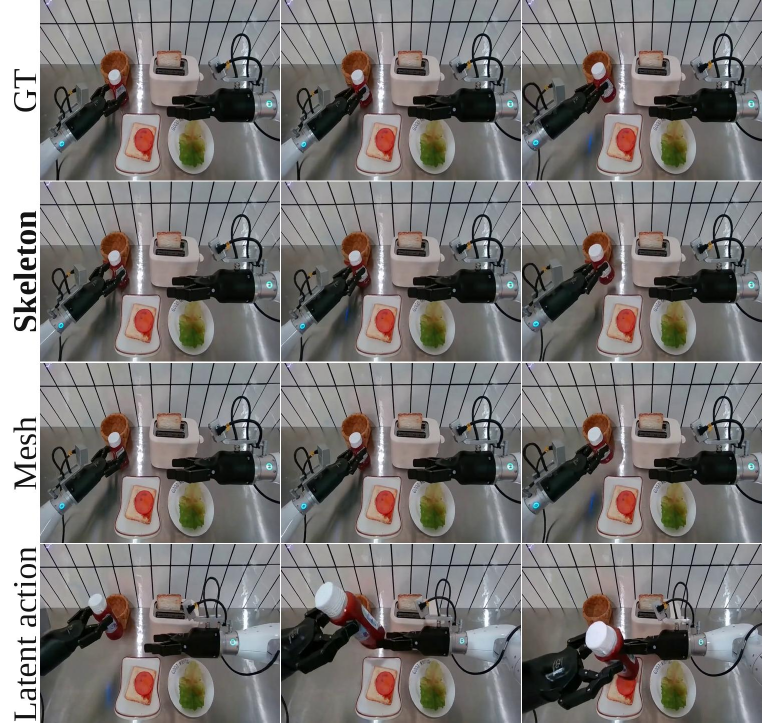} &
\includegraphics[width=0.49\linewidth]{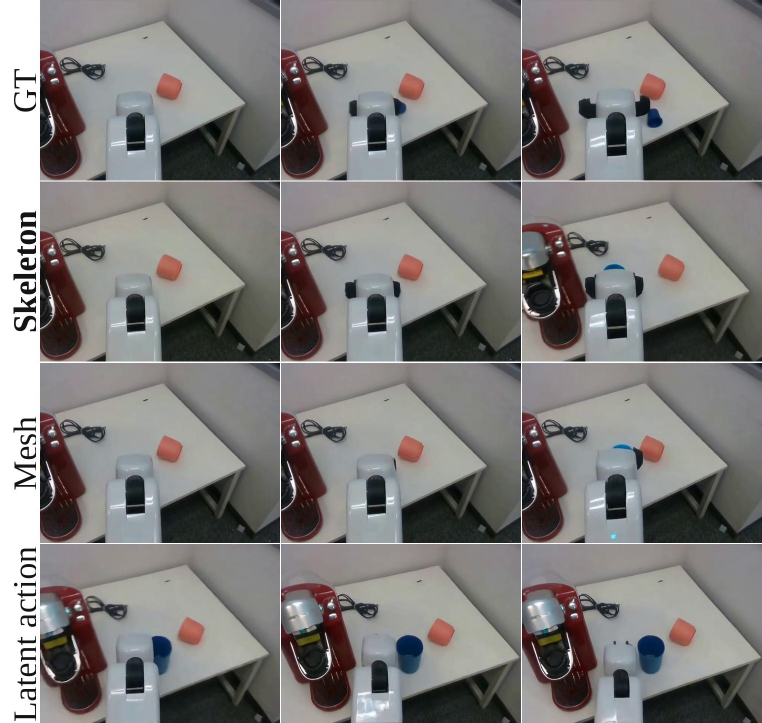} \\
\textbf{\textcolor{cAgiBot}{AgiBot G1}} &
\textbf{\textcolor{cAIROA}{AIROA-MoMa}} \\[2pt]
\includegraphics[width=0.49\linewidth]{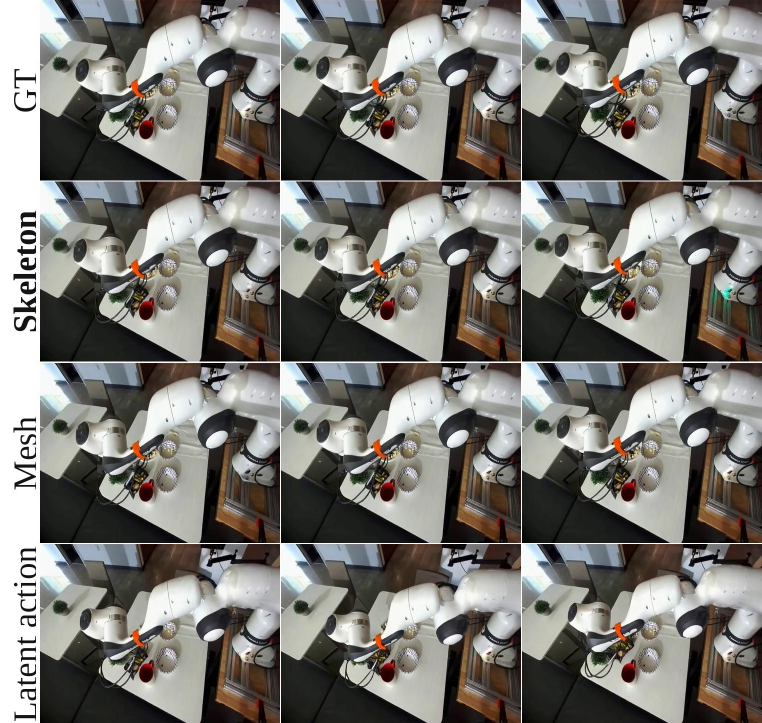} &
\includegraphics[width=0.49\linewidth]{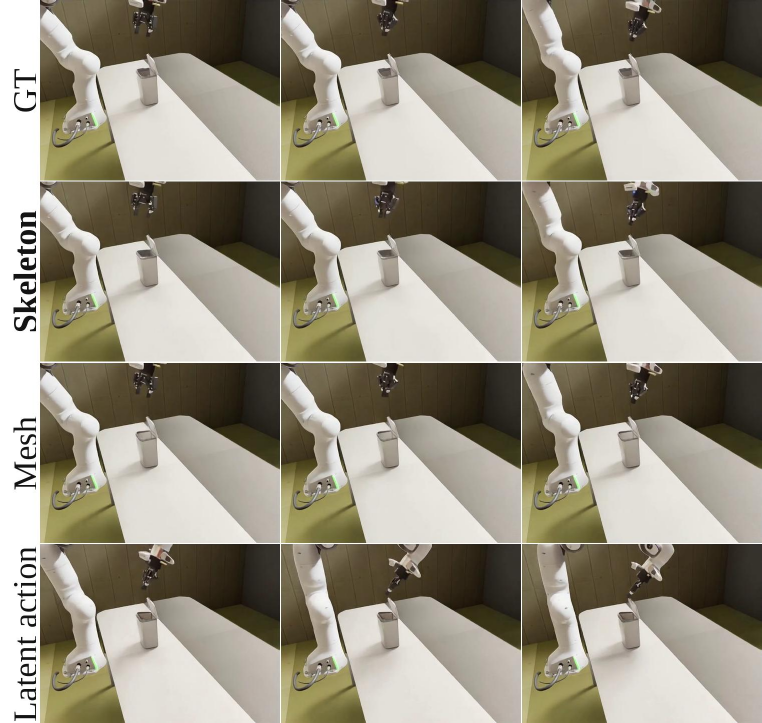} \\
\textbf{\textcolor{cDROID}{DROID}} &
\textbf{\textcolor{cIntern}{InternData}} \\[2pt]
\includegraphics[width=0.49\linewidth]{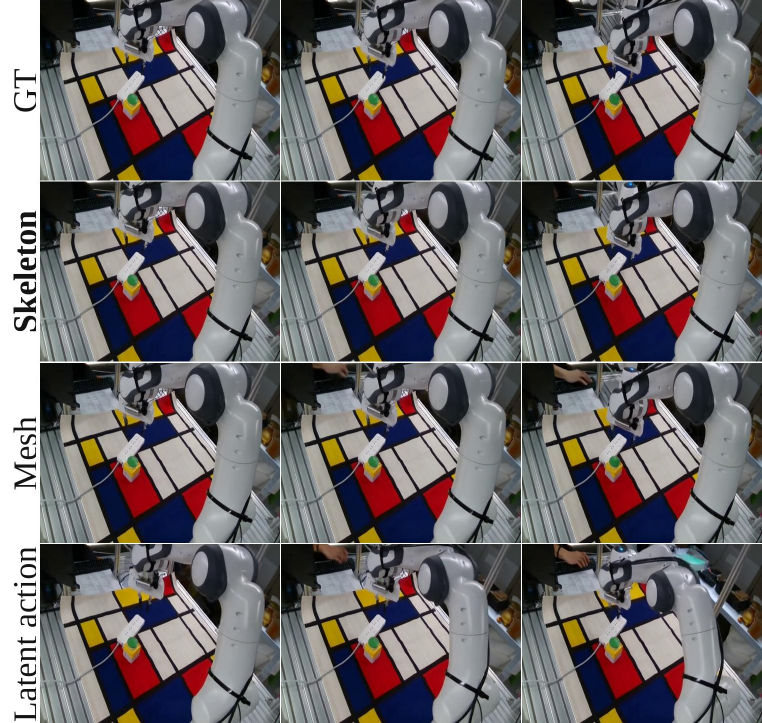} &
\includegraphics[width=0.49\linewidth]{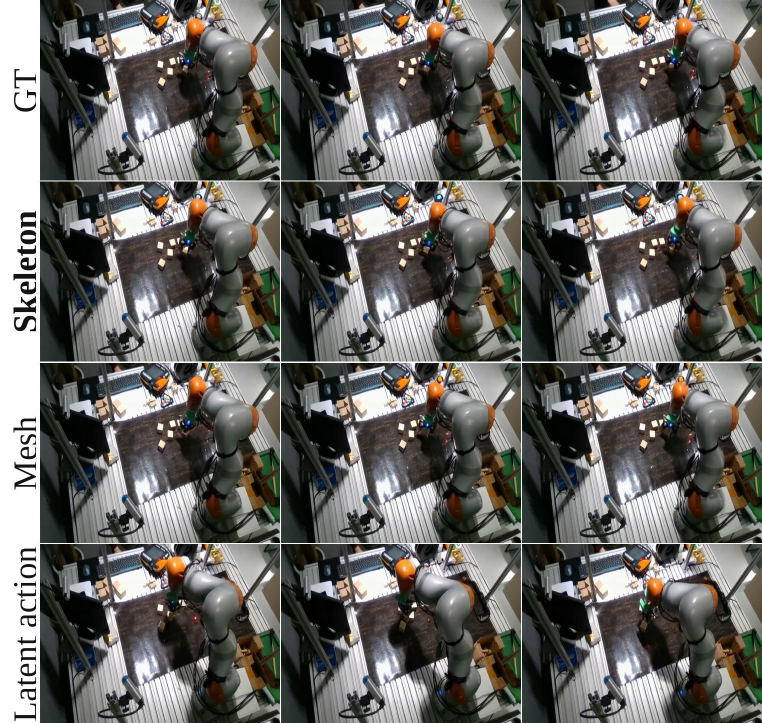} \\
\textbf{\textcolor{cRH20T5}{RH20T-cfg5}} &
\textbf{\textcolor{cRH20T7}{RH20T-cfg7}} \\
\end{tabular}
\caption{Conditioning-channel ablation (Table~\ref{tab:abl}, top block), one sample per embodiment. Rows: GT, skeleton (canonical, bold), URDF mesh render, latent action. Latent action consistently distorts gripper geometry, arm pose, and scene contents, while skeleton and mesh track ground truth comparably.}
\label{fig:qual_abl_cond}
\end{figure}

\begin{figure}[t]
\centering
\setlength{\tabcolsep}{2pt}
\renewcommand{\arraystretch}{1.0}
\begin{tabular}{@{}cc@{}}
\includegraphics[width=0.49\linewidth]{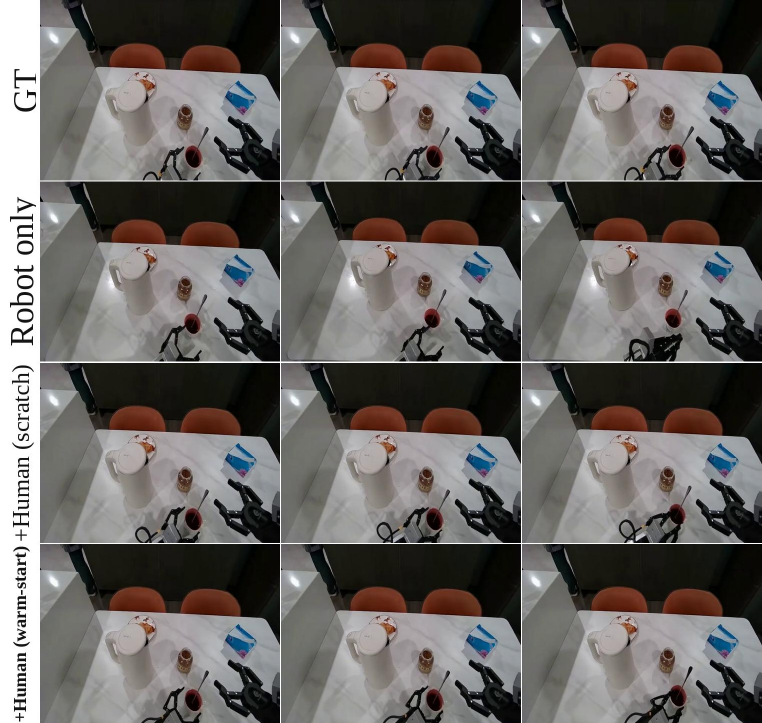} &
\includegraphics[width=0.49\linewidth]{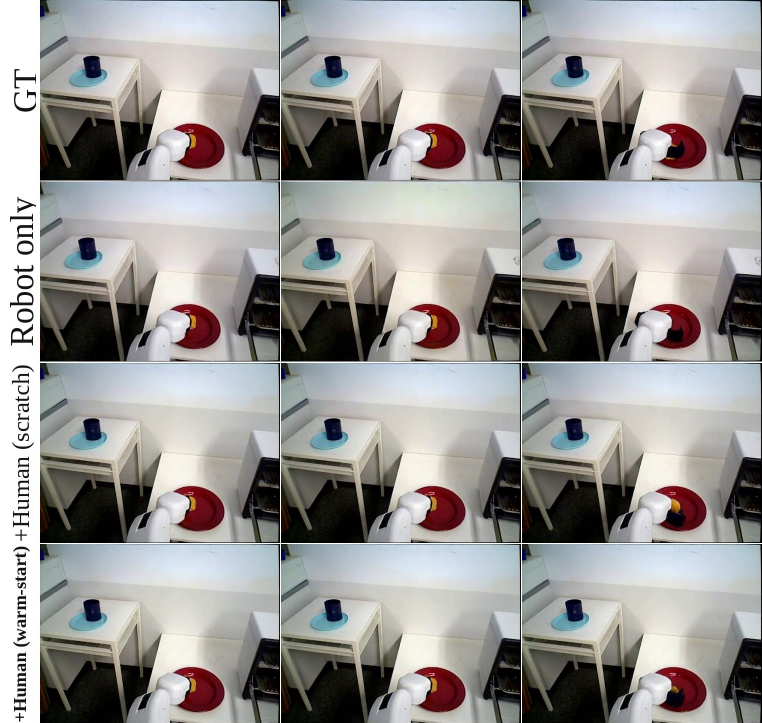} \\
\textbf{\textcolor{cAgiBot}{AgiBot G1}} &
\textbf{\textcolor{cAIROA}{AIROA-MoMa}} \\[2pt]
\includegraphics[width=0.49\linewidth]{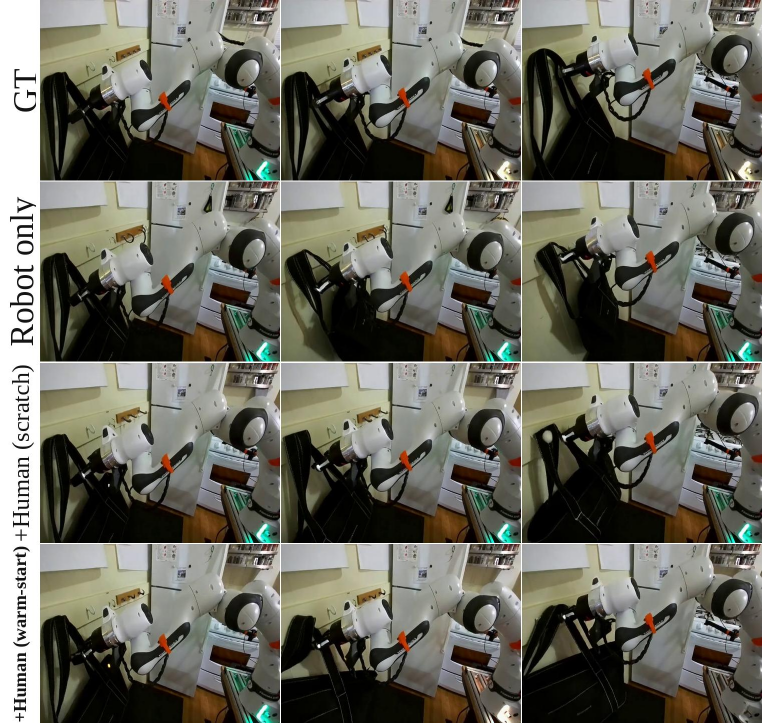} &
\includegraphics[width=0.49\linewidth]{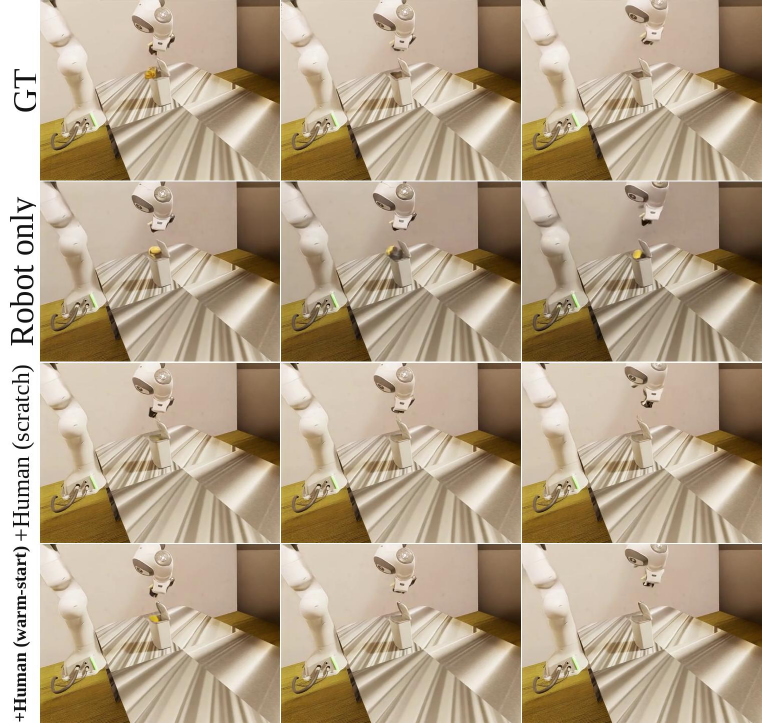} \\
\textbf{\textcolor{cDROID}{DROID}} &
\textbf{\textcolor{cIntern}{InternData}} \\[2pt]
\includegraphics[width=0.49\linewidth]{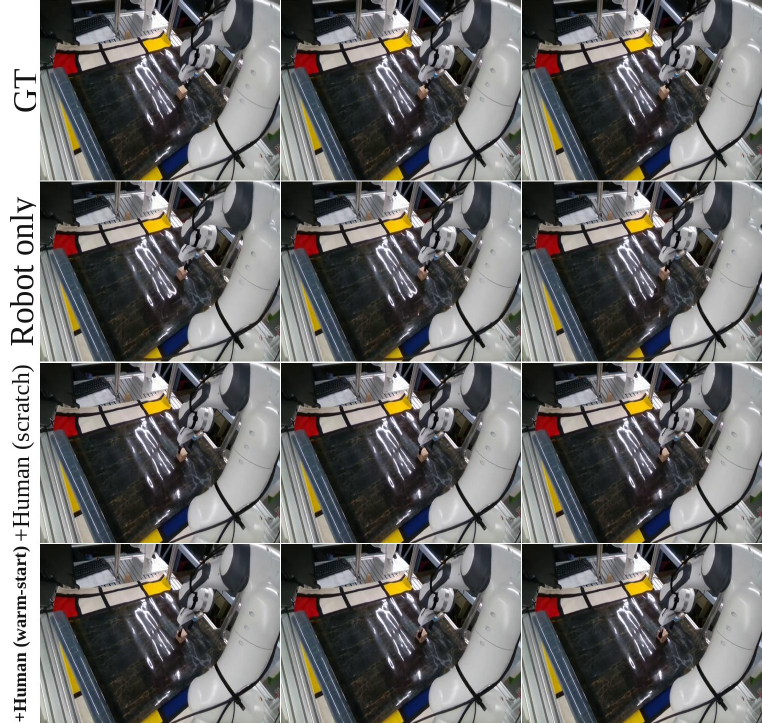} &
\includegraphics[width=0.49\linewidth]{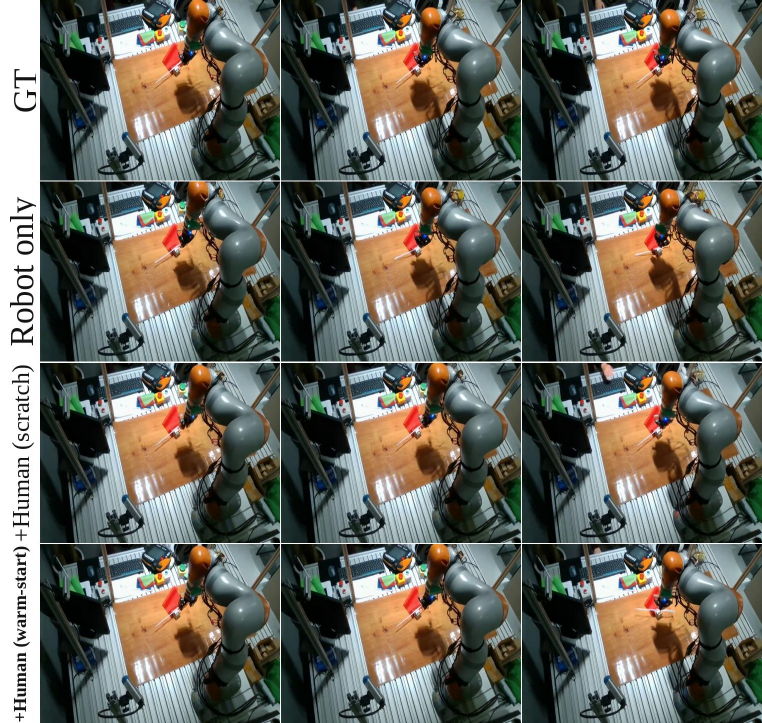} \\
\textbf{\textcolor{cRH20T5}{RH20T-cfg5}} &
\textbf{\textcolor{cRH20T7}{RH20T-cfg7}} \\
\end{tabular}
\caption{Data-composition ablation (Table~\ref{tab:abl}, bottom block), one sample per embodiment. Rows: GT, robot only, +human (train from scratch), +human (warm-start from robot only, the canonical bold row). Adding curated human clips and warm-starting consistently move predictions closer to ground truth across all six embodiments.}
\label{fig:qual_abl_data}
\end{figure}

We give more qualitative visualizations for the ablations in \S\ref{sec:exp_ablation}, covering all six robot embodiments: Figure~\ref{fig:qual_abl_cond} for the conditioning-channel ablation and Figure~\ref{fig:qual_abl_data} for data composition. Each panel stacks four rows at three time steps. Predictions appear without skeleton overlay so the comparison reflects output quality alone; row labels carry the conditioning identity. The bold row marks the canonical \methodname{} configuration.

\subsection{Human-data qualitative samples}
\label{app:human_qual}

\begin{figure}[t]
\centering
\setlength{\tabcolsep}{2pt}
\renewcommand{\arraystretch}{1.0}
\begin{tabular}{@{}cc@{}}
\includegraphics[width=0.49\linewidth]{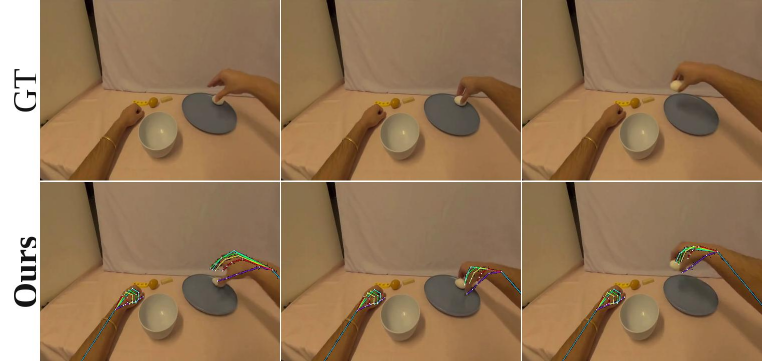} &
\includegraphics[width=0.49\linewidth]{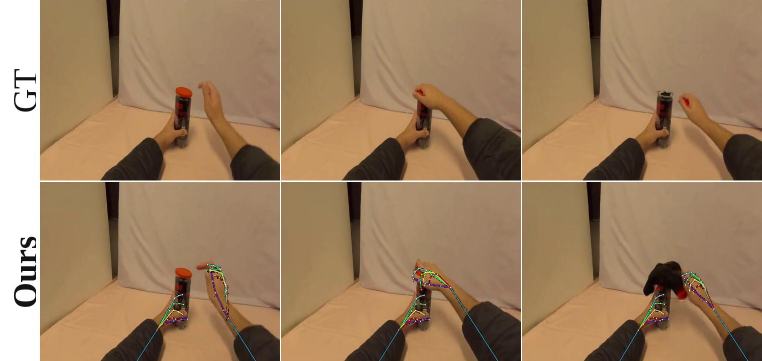} \\
\multicolumn{2}{c}{\textbf{\textcolor{cEgoDex}{EgoDex}} \footnotesize{(in-training)}} \\[2pt]
\includegraphics[width=0.49\linewidth]{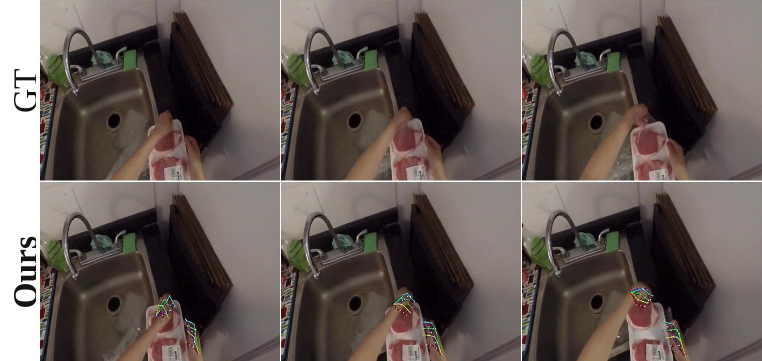} &
\includegraphics[width=0.49\linewidth]{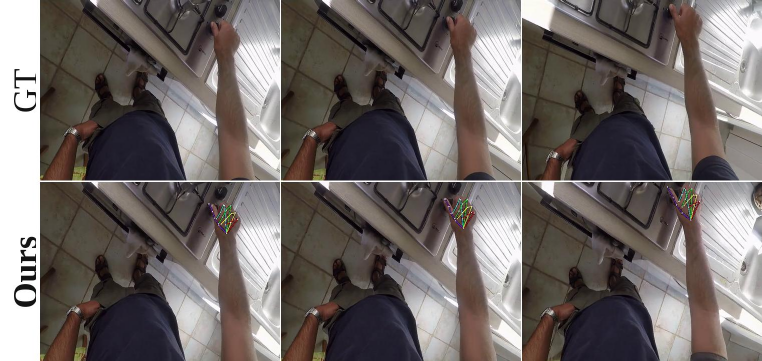} \\
\multicolumn{2}{c}{\textbf{\textcolor{cEPIC}{EPIC-Kitchens}} \footnotesize{(in-training)}} \\[2pt]
\includegraphics[width=0.49\linewidth]{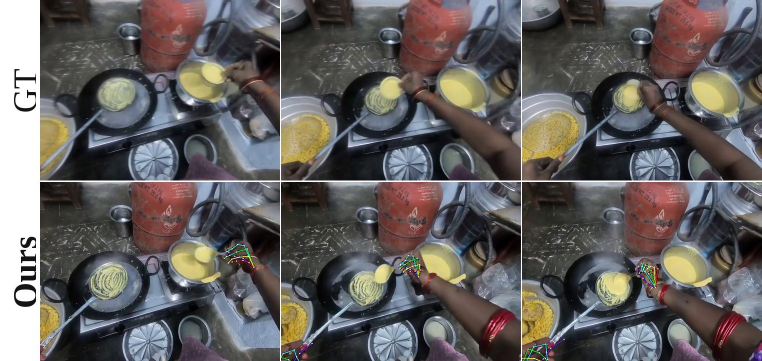} &
\includegraphics[width=0.49\linewidth]{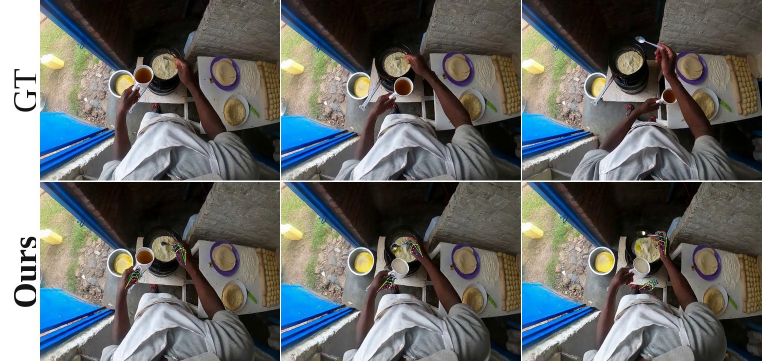} \\
\multicolumn{2}{c}{\textbf{\textcolor{cEgo4DCook}{Ego4D Cooking}} \footnotesize{(OOD)}} \\[2pt]
\includegraphics[width=0.49\linewidth]{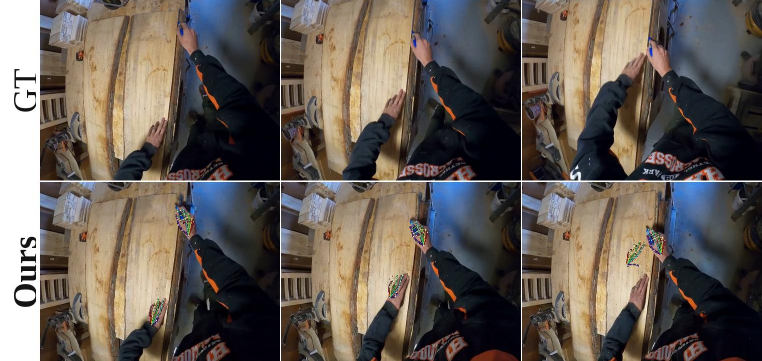} &
\includegraphics[width=0.49\linewidth]{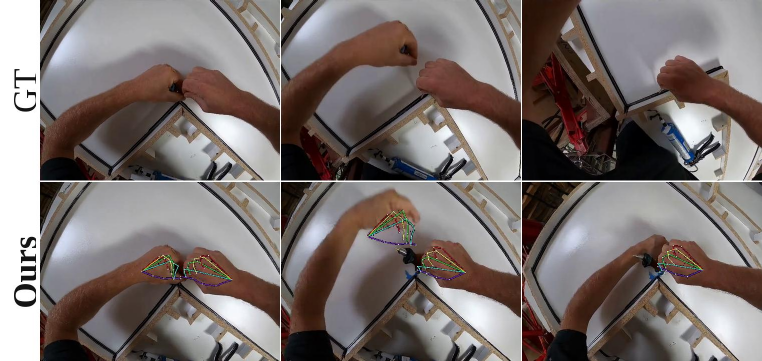} \\
\multicolumn{2}{c}{\textbf{\textcolor{cEgo4DOther}{Ego4D Other}} \footnotesize{(OOD)}} \\
\end{tabular}
\caption{Human-data qualitative samples. Each panel stacks GT (top) and \methodname{} (Ours, bold, with MANO skeleton overlay) at three time steps. Top two rows of panels are in-training datasets (EgoDex, EPIC-Kitchens); bottom two are out-of-distribution Ego4D subsets included for OOD probing only. The MANO conditioning constrains hand and arm motion across all four sources, even for OOD scenes where pixel-level fidelity is lower because of unseen environments.}
\label{fig:qual_human}
\end{figure}

\begin{figure}[t]
\centering
\setlength{\tabcolsep}{2pt}
\renewcommand{\arraystretch}{1.0}
\begin{tabular}{@{}cc@{}}
\includegraphics[width=0.49\linewidth]{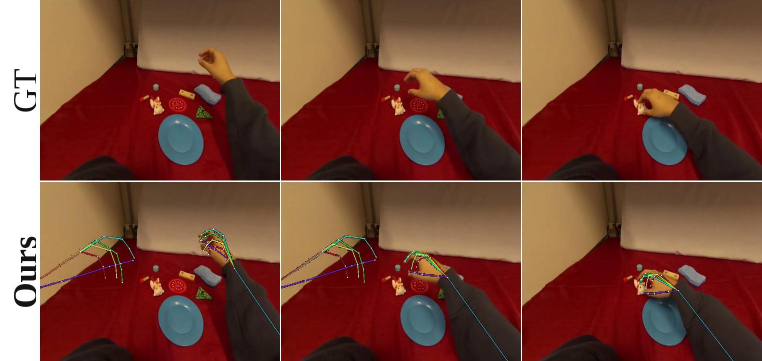} &
\includegraphics[width=0.49\linewidth]{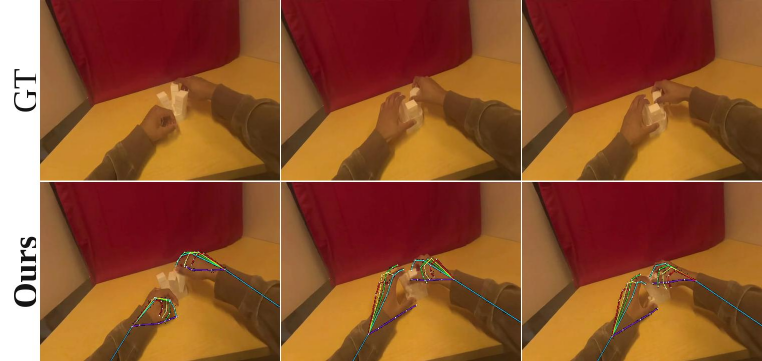} \\
\multicolumn{2}{c}{\textbf{\textcolor{cEgoDex}{EgoDex}} \footnotesize{(in-training)}} \\[2pt]
\includegraphics[width=0.49\linewidth]{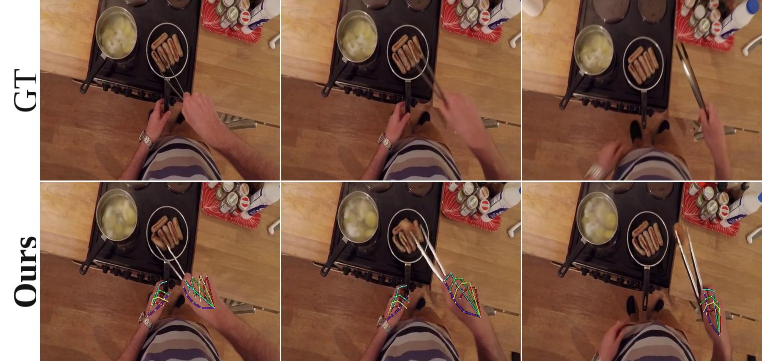} &
\includegraphics[width=0.49\linewidth]{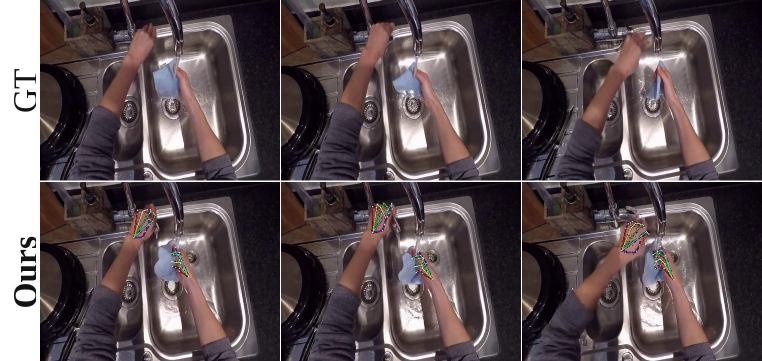} \\
\multicolumn{2}{c}{\textbf{\textcolor{cEPIC}{EPIC-Kitchens}} \footnotesize{(in-training)}} \\[2pt]
\includegraphics[width=0.49\linewidth]{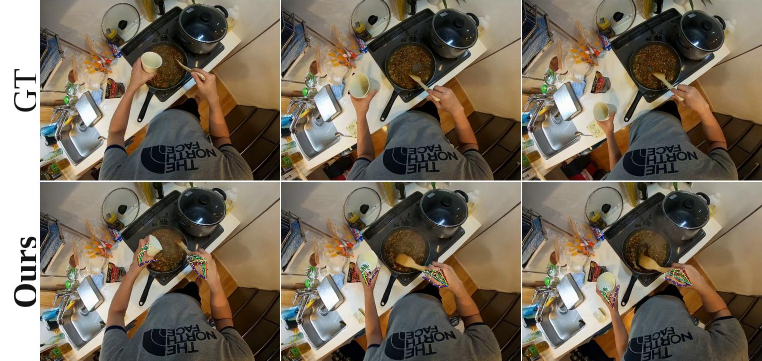} &
\includegraphics[width=0.49\linewidth]{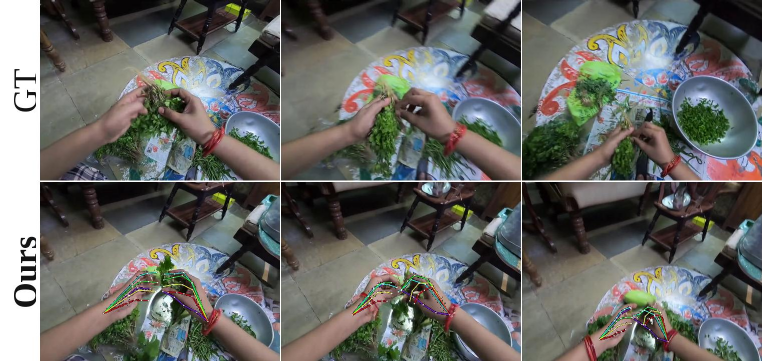} \\
\multicolumn{2}{c}{\textbf{\textcolor{cEgo4DCook}{Ego4D Cooking}} \footnotesize{(OOD)}} \\[2pt]
\includegraphics[width=0.49\linewidth]{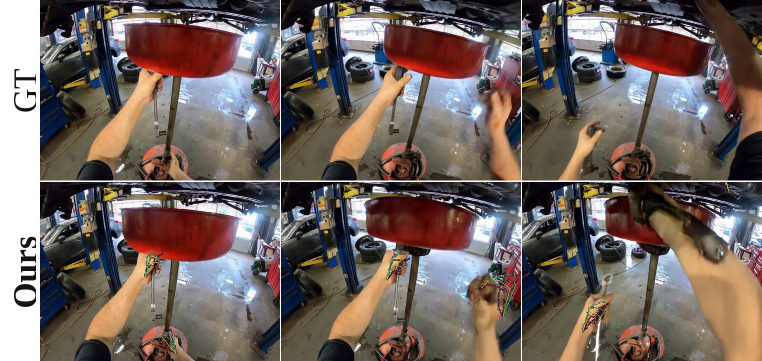} &
\includegraphics[width=0.49\linewidth]{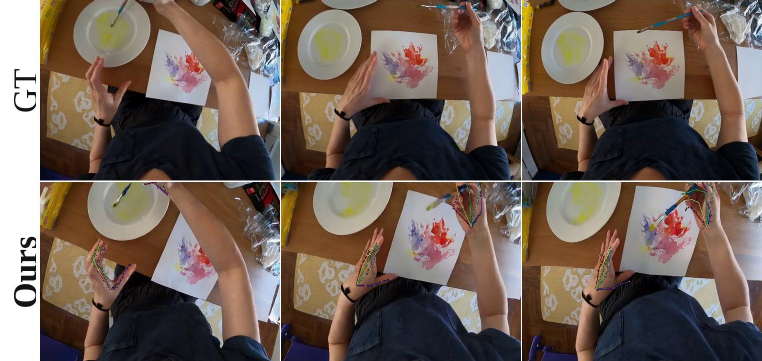} \\
\multicolumn{2}{c}{\textbf{\textcolor{cEgo4DOther}{Ego4D Other}} \footnotesize{(OOD)}} \\
\end{tabular}
\caption{Additional human-data qualitative samples, complementing Figure~\ref{fig:qual_human}. Same row layout as before. The Ego4D rows include outdoor and small-scene clips that are absent from EgoDex / EPIC-Kitchens training.}
\label{fig:qual_human_more}
\end{figure}

The main paper shows robot data generation only, but the same model also supports human scene generation: it accepts human MANO skeletons as conditioning inputs (\S\ref{sec:control}). Figures~\ref{fig:qual_human} and~\ref{fig:qual_human_more} show samples from four egocentric human datasets. EgoDex and EPIC-Kitchens are in-distribution, part of our training mixture; the cooking and non-cooking subsets of Ego4D~\citep{ego4d} are out-of-distribution probes held out from training. Each panel pairs ground truth with \methodname{} (Ours) at three time steps; the bold \emph{Ours} row shows the prediction with the conditioning MANO skeleton overlaid.

\subsection{Policy evaluation}
\label{app:policy_eval}

\subsubsection{Platform and camera calibration}
\label{app:roboarena_setup}

RoboArena~\citep{roboarena} instantiates each session on the DROID platform~\citep{droid}: a Franka Panda 7-DoF arm with a Robotiq 2F-85 parallel-jaw gripper, a ZED-mini stereo wrist camera, and one or more external ZED 2 stereo cameras. The dump releases synchronised RGB videos and joint trajectories but no per-session camera calibration; we therefore estimate the intrinsics with MoGe-v2~\citep{mogev2} and the static cam-to-base extrinsic with CtRNet-X~\citep{ctrnetx}, which regresses per-frame 2D keypoints and solves BPnP against URDF forward kinematics. We manually inspect the left-camera overlay quality and retain $65$ sessions, giving $65{\times}7{=}455$ (session, policy) cells.

\subsubsection{Metric definitions}
\label{app:policy_metrics}

We compare three conditioning channels from \S\ref{sec:exp_ablation}: skeleton against latent-action (global adaLN token from flattened end-effector chunks) and mesh (textured URDF render). Following SIMPLER~\citep{simpler} and WorldGym~\citep{worldgym}, we report four fidelity metrics. (i) The Mean Maximum Rank Violation $\mathrm{MMRV}$ on the seven-policy rank vector ($\downarrow$, range $[0, 6]$). (ii) Spearman $\rho$ between RoboArena and \methodname{} per-policy rank vectors ($\uparrow$). (iii) Pearson $r$ between RoboArena and \methodname{} per-policy mean binary success rates ($\uparrow$). (iv) The SR difference $\mathrm{SISR}_{\Delta} = |\mathrm{SR}_{\mathrm{real}} - \mathrm{SR}_{\mathrm{pred}}|$ in percentage points ($\downarrow$), the mean absolute error between real and predicted per-policy binary success rates.

\subsubsection{VLM success scoring}
\label{app:policy_eval_vlm}

We score each rollout with a vision-language model that stands in for a human RoboArena evaluator. For each rollout we sample 32 frames uniformly from the generated video at its native $512\times288$ resolution. These frames go to GPT-5 (\texttt{gpt-5-2025-08-07}) in temporal order, at high reasoning effort. The model sees only the task instruction and the frames, the evidence a human rater would see, and not the caption that conditions \methodname{}. It returns one JSON object: a binary success flag, a partial-progress score from 0 to 100, and a one-sentence reason. The box below gives the exact prompt.

\begin{tcolorbox}[
  breakable,
  colback=black!2,
  colframe=black!35,
  boxrule=0.4pt,
  arc=1mm,
  left=1mm,right=1mm,top=0.8mm,bottom=0.8mm,
  title={Prompt for the VLM success evaluator},
  fonttitle=\bfseries\footnotesize,
  fontupper=\ttfamily\footnotesize,
]
\raggedright\sloppy
You are evaluating whether a robot manipulation rollout successfully completes a described task. You see frames sampled uniformly across the rollout duration.

Output one JSON object only, no prose:

\{"binary": 0 or 1, "partial": 0-100, "reason": "\textless one short sentence\textgreater"\}

- binary = 1 only if the task is fully completed as described

- partial: 0 = no progress, 100 = fully complete, intermediate = partial progress

Task: \textless task instruction\textgreater

Frames in temporal order: \textless 32 images\textgreater
\end{tcolorbox}

To check that the model agrees with people, we calibrate it against real RoboArena videos that carry human labels. We draw 100 real-robot clips, balanced across 50 successes and 50 failures, and score them with the same prompt. The model matches the human binary label on 78 of 100 clips. It rarely calls a failure a success (specificity 0.90), and it misses about a third of the real successes (recall 0.66), so it under-reports success rather than inflating it. On this balanced set the VLM binary verdict agrees with the human binary label well above chance (Pearson $r=0.58$, $p<10^{-7}$).

\subsubsection{Qualitative comparison}
\label{app:policy_eval_qual}

\begin{figure}[p]
\centering
\footnotesize
\setlength{\tabcolsep}{3pt}
\renewcommand{\arraystretch}{1.1}
\begin{tabular}{@{}>{\raggedright\arraybackslash}m{3.3cm} m{0.55cm} m{0.55cm} m{10.4cm}@{}}
\toprule
Policy & real & WM & \multicolumn{1}{c}{\scriptsize time $\rightarrow$} \\
\midrule
$\pi_0$-flow-DROID & \xmark & \xmark & \includegraphics[width=10.4cm]{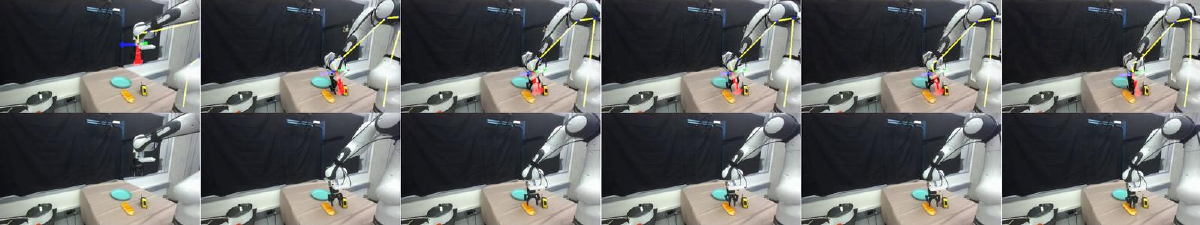} \\
$\pi_0$-FAST-DROID & \cmark & \cmark & \includegraphics[width=10.4cm]{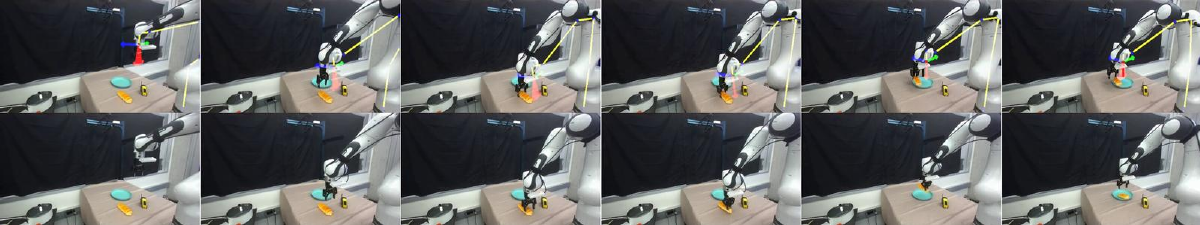} \\
PG-flow-DROID & \cmark & \cmark & \includegraphics[width=10.4cm]{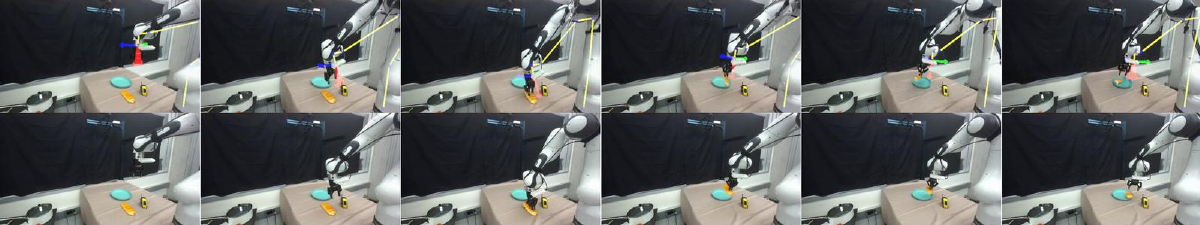} \\
PG-FSQ-DROID & \cmark & \cmark & \includegraphics[width=10.4cm]{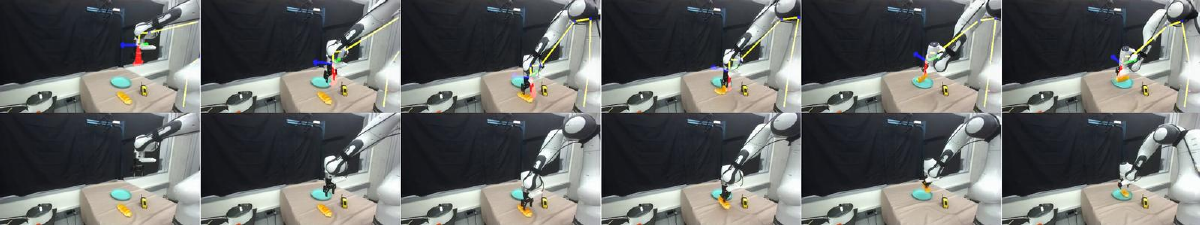} \\
PG-FAST-DROID & \cmark & \cellcolor{red!12}\xmark & \includegraphics[width=10.4cm]{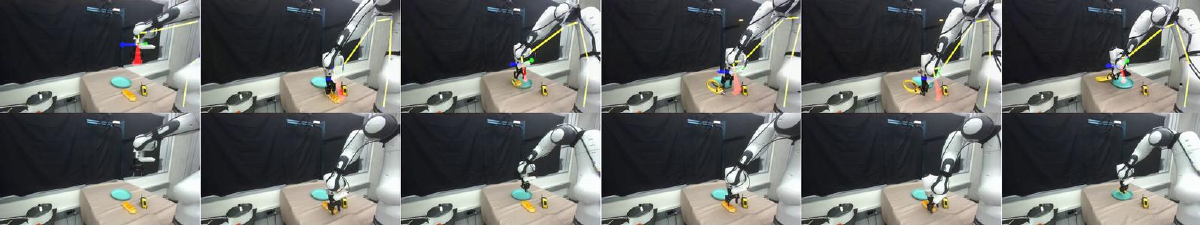} \\
PG-FAST+-DROID & \cmark & \cmark & \includegraphics[width=10.4cm]{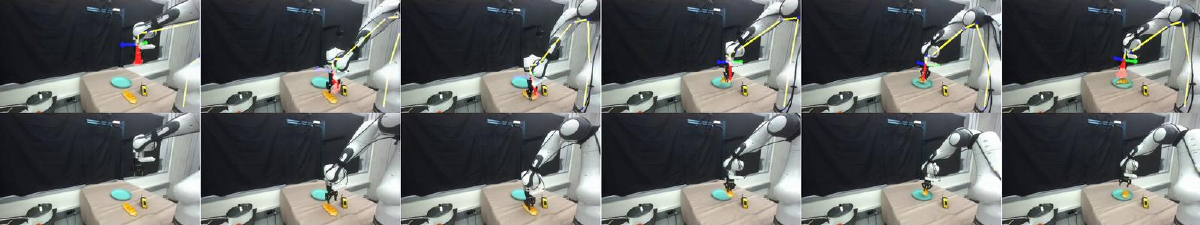} \\
PG-Bin-DROID & \xmark & \xmark & \includegraphics[width=10.4cm]{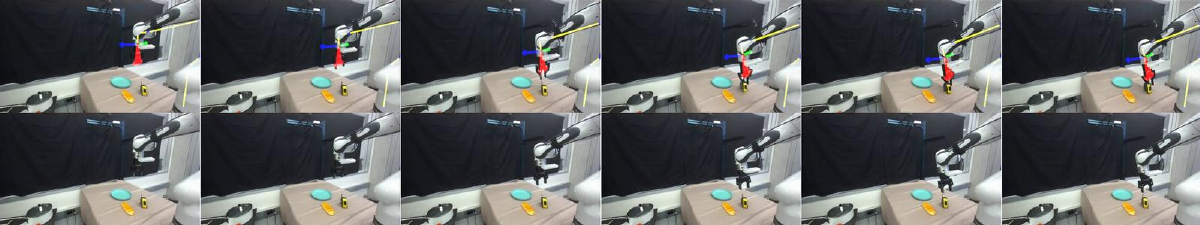} \\
\bottomrule
\end{tabular}
\caption{\textbf{Put the food on the plate.} Each strip pairs the \methodname{} rollout (top) with the real-robot video (bottom) at six time steps. The \emph{real} column is the RoboArena human success label and the \emph{WM} column is the VLM verdict on the rollout (\S\ref{sec:exp_policy_eval}); cells where the two differ are highlighted. On this task 5 of seven policies succeed on the real robot.}
\label{fig:policy_eval_81a85b7c}
\end{figure}

\begin{figure}[p]
\centering
\footnotesize
\setlength{\tabcolsep}{3pt}
\renewcommand{\arraystretch}{1.1}
\begin{tabular}{@{}>{\raggedright\arraybackslash}m{3.3cm} m{0.55cm} m{0.55cm} m{10.4cm}@{}}
\toprule
Policy & real & WM & \multicolumn{1}{c}{\scriptsize time $\rightarrow$} \\
\midrule
$\pi_0$-flow-DROID & \cmark & \cellcolor{red!12}\xmark & \includegraphics[width=10.4cm]{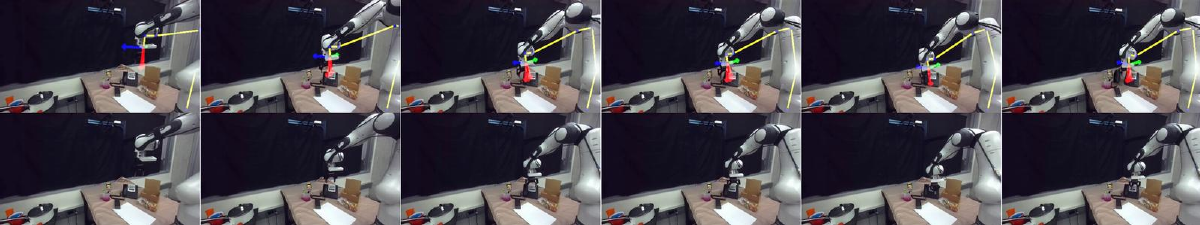} \\
$\pi_0$-FAST-DROID & \xmark & \xmark & \includegraphics[width=10.4cm]{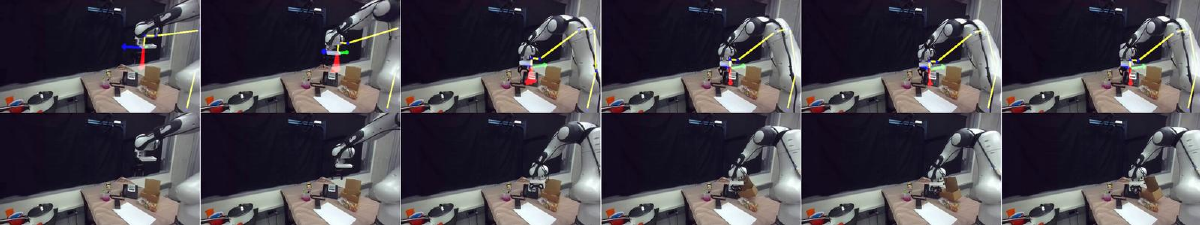} \\
PG-flow-DROID & \xmark & \xmark & \includegraphics[width=10.4cm]{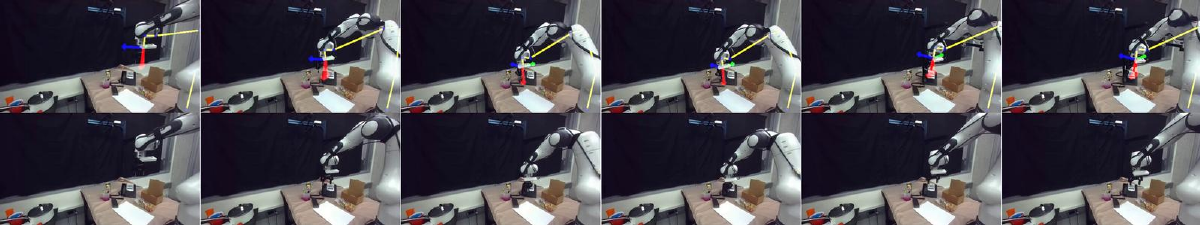} \\
PG-FSQ-DROID & \xmark & \xmark & \includegraphics[width=10.4cm]{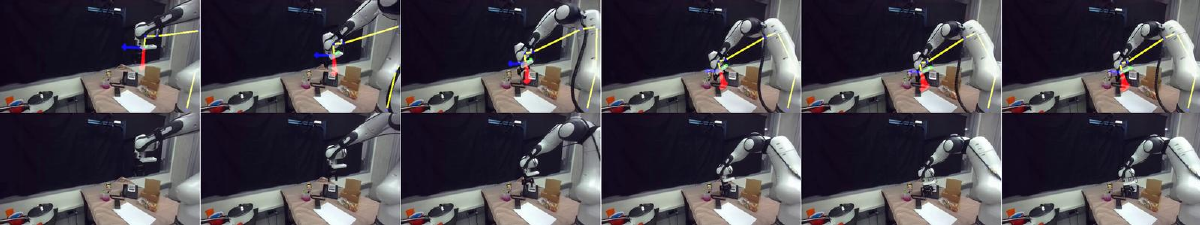} \\
PG-FAST-DROID & \cmark & \cmark & \includegraphics[width=10.4cm]{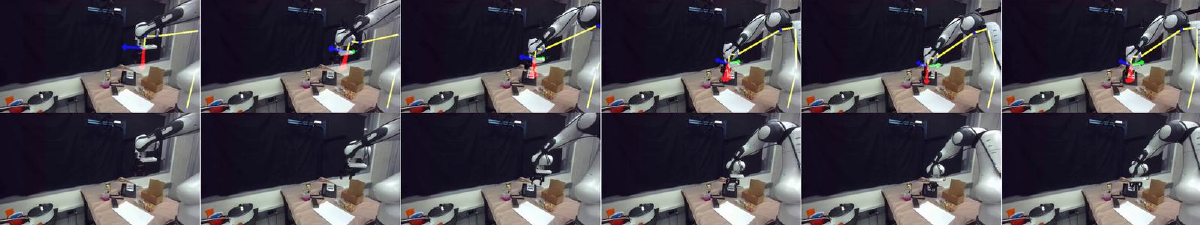} \\
PG-FAST+-DROID & \xmark & \xmark & \includegraphics[width=10.4cm]{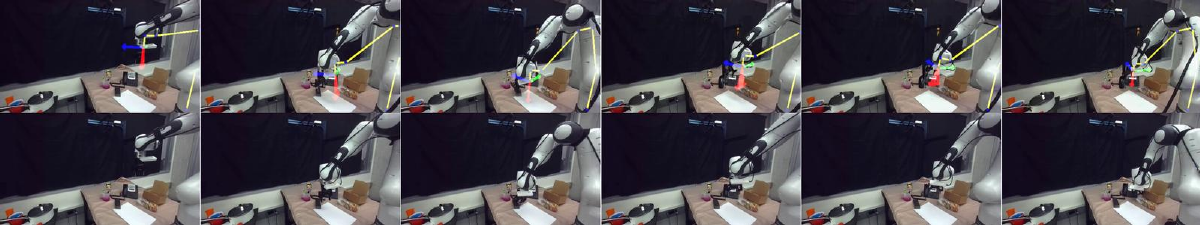} \\
PG-Bin-DROID & \xmark & \xmark & \includegraphics[width=10.4cm]{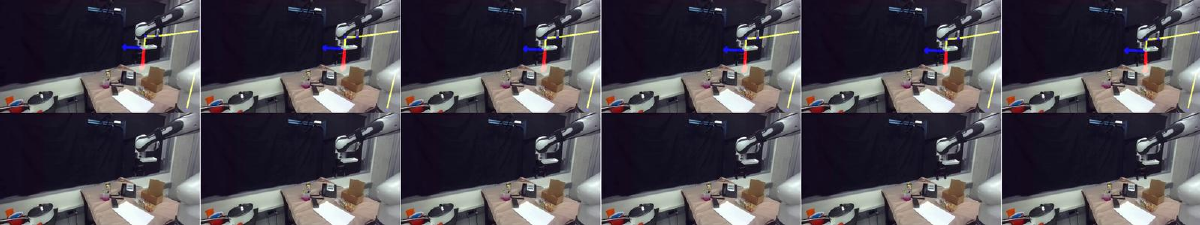} \\
\bottomrule
\end{tabular}
\caption{\textbf{Press a button on the phone.} Each strip pairs the \methodname{} rollout (top) with the real-robot video (bottom) at six time steps. The \emph{real} column is the RoboArena human success label and the \emph{WM} column is the VLM verdict on the rollout (\S\ref{sec:exp_policy_eval}); cells where the two differ are highlighted. On this task 2 of seven policies succeed on the real robot.}
\label{fig:policy_eval_7ac4ded2}
\end{figure}

\begin{figure}[p]
\centering
\footnotesize
\setlength{\tabcolsep}{3pt}
\renewcommand{\arraystretch}{1.1}
\begin{tabular}{@{}>{\raggedright\arraybackslash}m{3.3cm} m{0.55cm} m{0.55cm} m{10.4cm}@{}}
\toprule
Policy & real & WM & \multicolumn{1}{c}{\scriptsize time $\rightarrow$} \\
\midrule
$\pi_0$-flow-DROID & \xmark & \xmark & \includegraphics[width=10.4cm]{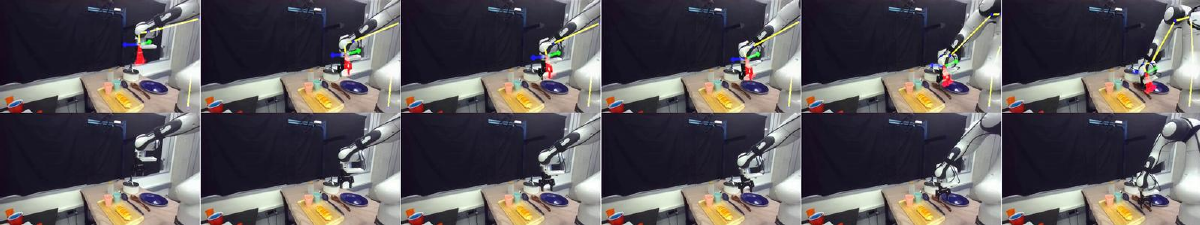} \\
$\pi_0$-FAST-DROID & \xmark & \xmark & \includegraphics[width=10.4cm]{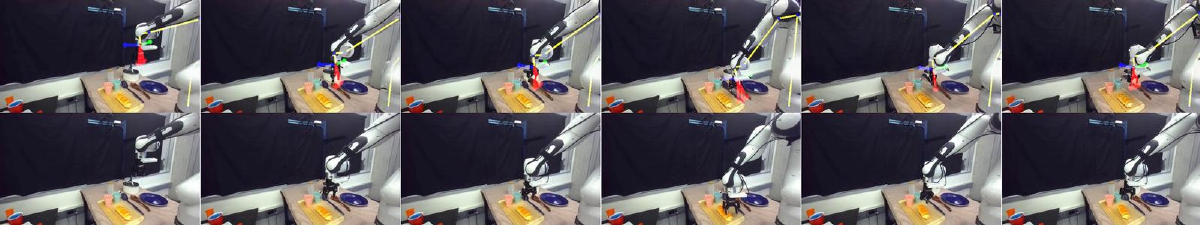} \\
PG-flow-DROID & \xmark & \xmark & \includegraphics[width=10.4cm]{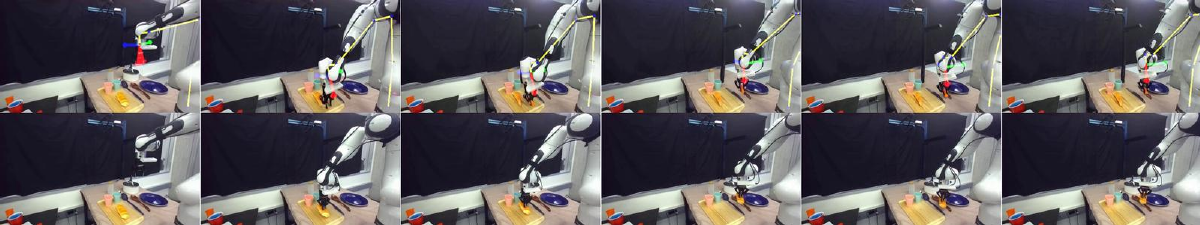} \\
PG-FSQ-DROID & \xmark & \xmark & \includegraphics[width=10.4cm]{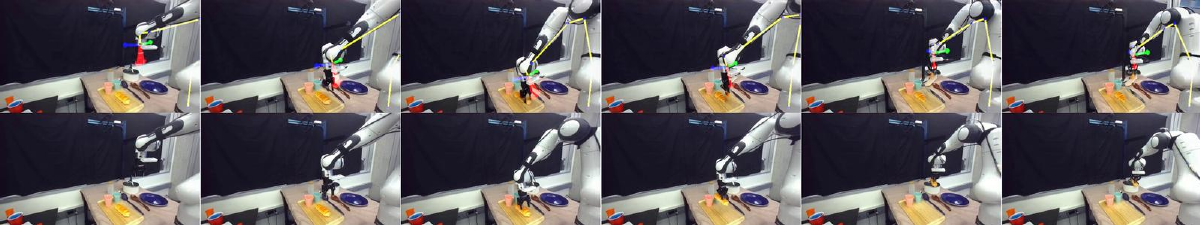} \\
PG-FAST-DROID & \xmark & \xmark & \includegraphics[width=10.4cm]{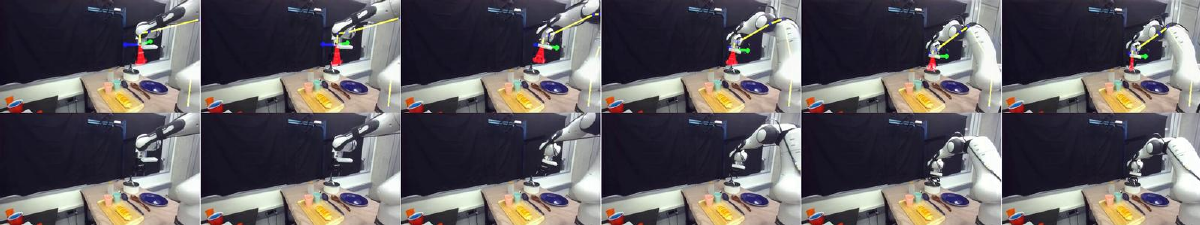} \\
PG-FAST+-DROID & \cmark & \cmark & \includegraphics[width=10.4cm]{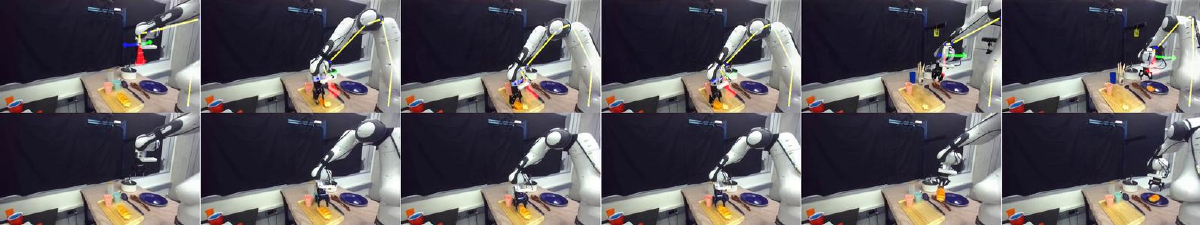} \\
PG-Bin-DROID & \xmark & \xmark & \includegraphics[width=10.4cm]{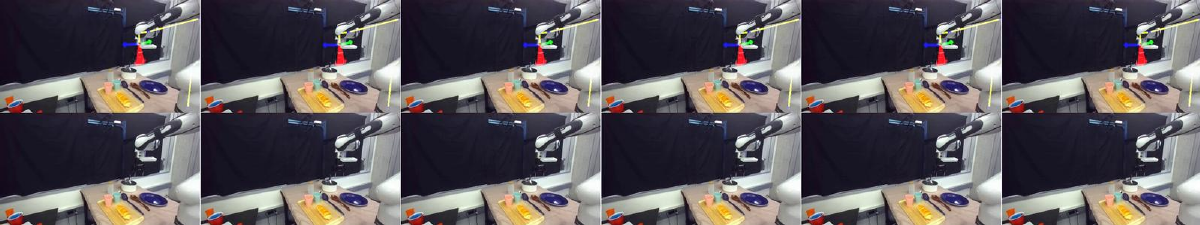} \\
\bottomrule
\end{tabular}
\caption{\textbf{Move the bread to the plate.} Each strip pairs the \methodname{} rollout (top) with the real-robot video (bottom) at six time steps. The \emph{real} column is the RoboArena human success label and the \emph{WM} column is the VLM verdict on the rollout (\S\ref{sec:exp_policy_eval}); cells where the two differ are highlighted. On this task 1 of seven policies succeeds on the real robot.}
\label{fig:policy_eval_4ba7c1e8}
\end{figure}

Figures~\ref{fig:policy_eval_81a85b7c}, \ref{fig:policy_eval_7ac4ded2}, and~\ref{fig:policy_eval_4ba7c1e8} show three typical tasks, one policy per row. For each session we roll out all seven DROID policies from the recorded first frame and pair each rollout with the matching real-robot video. The top band of a strip is the \methodname{} rollout and the bottom band is the real-robot video, at six time steps. Frame by frame, the rollout follows the real arm and the objects it touches. The \emph{real} and \emph{WM} columns give the RoboArena human label and the VLM verdict, and we highlight the cells where the two disagree.

The disagreements match the calibration above: the VLM is conservative. In these panels it records a false negative on a partial attempt, where the rollout shows the arm reaching and moving the target, the human evaluator credits partial progress, and the VLM still scores a failure (for example PG-FAST-DROID on \emph{put the food on the plate}). The rank and success-rate metrics over all 455 cells are in \S\ref{sec:exp_policy_eval}.

\subsection{Asset licenses}
\label{app:licenses}
\begin{table}[t]
  \centering
  \caption{Licenses of external assets used in this work.}
  \label{tab:asset_licenses}
  \scriptsize
  \setlength{\tabcolsep}{3pt}
  \renewcommand{\arraystretch}{1.05}
  \begin{tabular}{@{}p{0.40\linewidth}p{0.18\linewidth}p{0.38\linewidth}@{}}
    \toprule
    Asset & Type & License \\
    \midrule
    Cosmos-Predict2.5-2B~\citep{cosmos-predict2p5} & Base model & NVIDIA Open Model License; Apache 2.0 (code) \\
    DROID~\citep{droid} & Dataset & CC-BY 4.0 \\
    RH20T (cfg5, cfg7)~\citep{rh20t} & Dataset & CC-BY-SA 4.0 \\
    InternData-A1~\citep{interndataa1} & Dataset & CC-BY-NC-SA 4.0 \\
    AgiBot World~\citep{agibotworld} & Dataset & CC-BY-NC-SA 4.0 \\
    AIROA-MoMa~\citep{airoamoma} & Dataset & CC-BY-NC-SA 4.0 \\
    EgoDex~\citep{egodex} & Dataset & CC-BY-NC-ND 4.0 \\
    EPIC-Kitchens~\citep{damen2020epic} & Dataset & CC-BY-NC 4.0 \\
    VITRA~\citep{vitra} & Code/pipeline & MIT \\
    HaMeR & Code/model & Research / non-commercial \\
    MANO & Model & Research / non-commercial \\
    \bottomrule
  \end{tabular}
  \vspace{-\intextsep}
\end{table}

Table~\ref{tab:asset_licenses} summarizes the third-party assets used in this work and their corresponding licenses. 


\end{document}